\documentclass[10pt,twocolumn,letterpaper]{article}

\usepackage[pagenumbers]{cvpr} 

\usepackage{amsfonts,bm}
\usepackage[normalem]{ulem}
\usepackage{pifont}
\usepackage{float}
\useunder{\uline}{\ul}{}

\usepackage{amsmath,amsfonts,bm}

\def\eqref#1{equation~\ref{#1}}

\def\1{\bm{1}}

\DeclareMathAlphabet{\mathsfit}{\encodingdefault}{\sfdefault}{m}{sl}
\SetMathAlphabet{\mathsfit}{bold}{\encodingdefault}{\sfdefault}{bx}{n}

\definecolor{cvprblue}{rgb}{0.21,0.49,0.74}
\usepackage[pagebackref,breaklinks,colorlinks,allcolors=cvprblue]{hyperref}

\usepackage{multirow}
\usepackage[table,xcdraw]{xcolor}
\usepackage{tabularx}

\def\confName{CVPR}
\def\confYear{2027}

\title{ToWo-VTON: Take-Off Wear-On Virtual Try-On via Topology-aware Decoupling}

\author{
Zhenchen Wan$^{1,*}$ \quad
Tianxi Chen$^{2,*}$ \quad
Runqi Lin$^{3}$ \quad
Tianyi Bo$^{2}$ \quad
Ce Chen$^{2}$ \\
Jiaxin Huang$^{4}$ \quad
Yanwu Xu$^{5}$ \quad
Tongliang Liu$^{1,\dagger}$ \quad
Mingming Gong$^{2,\dagger}$ \\[6pt]
{\small
$^{1}$University of Sydney \quad
$^{2}$University of Melbourne \quad
$^{3}$University of Oxford
} \\
{\small
$^{4}$MBZUAI \quad
$^{5}$Google
} \\[4pt]
{\ttfamily\small
\{zwan0681, tongliang.liu\}@sydney.edu.au
} \\
{\ttfamily\small
tianxic@student.unimelb.edu.au \quad
mingming.gong@unimelb.edu.au
} \\[4pt]
{\small
$^{*}$ Equal contribution \qquad
$^{\dagger}$ Corresponding authors
}
}

\begin{document}

\twocolumn[{%
\renewcommand\twocolumn[1][]{#1}%
\maketitle
\begin{center}
    \centering
    \includegraphics[width=1\textwidth]{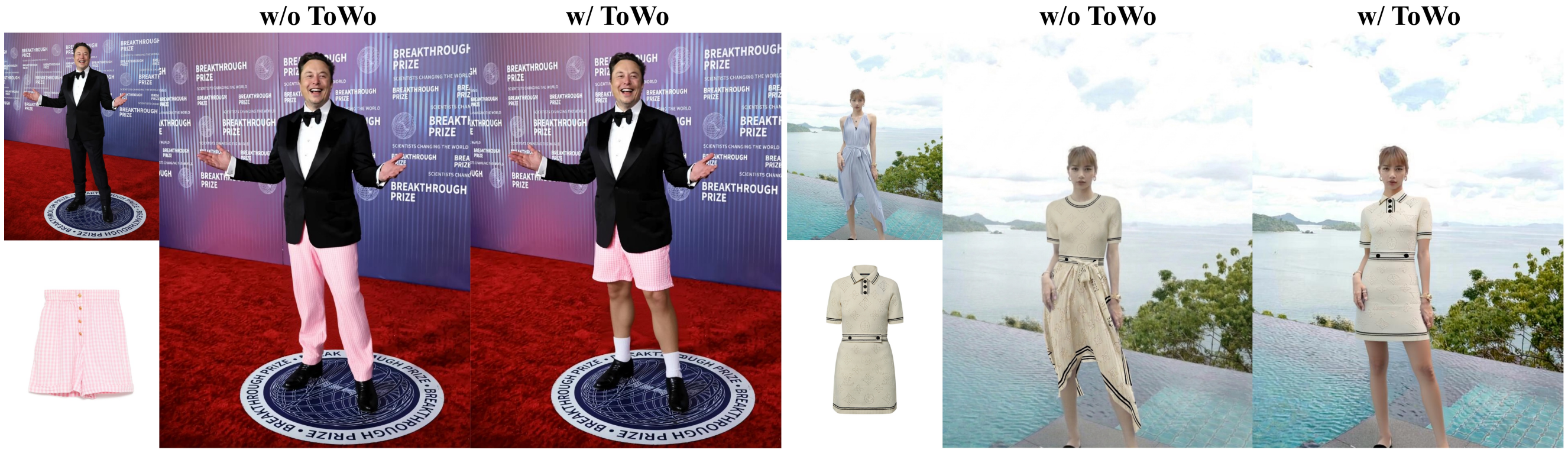}
    \captionsetup{type=figure, skip=-1pt}
    \captionof{figure}{\textbf{Effect of ToWo-VTON under large garment transformations.}
    Large length and category changes, such as long pants to shorts (left) or a long asymmetrical dress to a short polo dress (right), often cause severe topology artifacts without ToWo. 
    \textit{w/ ToWo} denotes Wear-On trained on $\mathcal{D}_{\text{BFDC}}$ and inferred with the two-stage Take-Off/Wear-On pipeline.}
    \label{fig:cover_image}
\end{center}%
}]

\begin{abstract}

Virtual Try-On (VTON) aims to synthesize garment transfer without physical trials. Although recent methods perform well on upper-body garments, they still struggle in lower-body and full-body scenarios with complex occlusions and large silhouette changes. To address this limitation, we introduce ToWo (Take-Off Wear-On), a two-stage paradigm that decouples body structure estimation from garment synthesis. Take-Off first generates a tight-fitting body proxy that provides estimated structural guidance, and Wear-On then synthesizes the target garment conditioned on this prior. This decoupling reduces source-garment interference, supporting garment placement under substantial silhouette variations. Built on this paradigm, we propose ToWo-VTON, a unified topology-aware framework with structure-aware supervision. We construct the Body-Fitting DressCode (BFDC) dataset via body-aware diffusion and VLM-based filtering to provide body-proxy supervision across upper-, lower-, and full-body cases. We instantiate ToWo-VTON on both diffusion-based and MLLM-guided backbones. For diffusion-based models, we fine-tune on BFDC using the body proxy as the person input. For the MLLM-guided model, we use native backbone conditioning for the body proxy and target garment, and add Garment Boosting Attention (GBA) to preserve garment details. Experiments on VITON-HD and DressCode demonstrate improvements over the evaluated baselines, with KID reductions of 26.1\% and 58.6\% for diffusion-based and MLLM-guided implementations, respectively. The combined BFDC-training and Take-Off/Wear-On strategy also produces more coherent limb layouts and garment boundaries across the evaluated garment categories.

\end{abstract}

\section{Introduction}
\label{sec:intro}

Virtual Try-On (VTON) has emerged as a promising technology in fashion and e-commerce, enabling realistic garment previews without requiring physical trials and supporting personalized online shopping~\citep{Han_2018_CVPR, Kim_2024_CVPR, 10.1007/978-3-031-73016-0_13, chong_catvton_2025, Yang_2024_CVPR, Kim_2025_ICCV}. Early methods relied on Generative Adversarial Networks (GANs)~\citep{goodfellow2020generative}, which often struggled with visual fidelity and garment diversity. More recently, diffusion-based VTON methods have improved realism and garment alignment, achieving strong results on upper-body benchmarks~\citep{Kim_2024_CVPR, 10.1007/978-3-031-73016-0_13}.

However, large source-to-target silhouette changes remain challenging, particularly in lower-body and full-body try-on. When the target garment differs substantially in length or coverage, the source garment can dominate the apparent contours and boundaries that guide synthesis, resulting in residual source-clothing patterns, distorted limbs, or misplaced garment boundaries. Such interference can persist through garment-shaped masks in mask-based systems and through the person image itself in mask-free systems. These observations motivate conditioning garment synthesis on a representation that makes body structure more accessible under clothing occlusions. Here, we use body topology to describe the spatial arrangement of limbs, body contours, and their relationship to the garment.

To address this limitation, we introduce ToWo (Take-Off Wear-On), a two-stage paradigm that decouples body structure estimation from garment synthesis. Take-Off first generates a tight-fitting body proxy that reduces source-garment occlusion and provides estimated guidance for pose, limb layout, and body contours. Wear-On then synthesizes the target garment conditioned on this structural prior. The image-based proxy supplies dense contours and spatial context in a form compatible with existing person-conditioning branches. This decoupling reduces source-garment interference and supports garment placement under substantial silhouette variations. Built on this paradigm, we propose ToWo-VTON, a unified topology-aware framework with structure-aware supervision.

To align training with body-proxy conditioning at inference, we construct the Body-Fitting DressCode (BFDC) dataset through body-aware diffusion and vision-language filtering. We develop the Body Structure Revealing (BSR) model by fine-tuning a diffusion backbone on a curated tight-fit subset, then use it to generate body proxies. A vision-language model filters the samples for visible artifacts and structural consistency. BFDC pairs the retained proxies with the original person and garment images, providing body-proxy supervision across upper-, lower-, and full-body cases. This allows Wear-On to learn the same type of structural conditioning supplied by Take-Off at inference.

We instantiate ToWo-VTON on both diffusion-based and MLLM-guided backbones. For mask-based and mask-free diffusion VTON models, we replace the original person input with the body proxy and fine-tune on BFDC using their existing architectures. For the MLLM-guided implementation, we supply the body proxy and target garment through OmniGen2's native reference-image conditioning pathway~\citep{wu2025omnigen2explorationadvancedmultimodal}. We further introduce Garment Boosting Attention (GBA) to inject garment features during denoising and preserve fine-grained garment details.

Experiments on VITON-HD~\citep{Choi_2021_CVPR} and DressCode~\citep{10.1007/978-3-031-20074-8_20} demonstrate improvements over the evaluated baselines. On DressCode Lower-body, Ours (BooW) reduces KID by 26.1\% relative to reproduced BooW-VTON, while the full MLLM-guided model reduces KID by 58.6\% relative to its direct Wear-On baseline (Ours w/o ToWo). These comparisons assess the combined BFDC-training and Take-Off/Wear-On strategy. Qualitative comparisons illustrate more coherent limb layouts and garment boundaries under large silhouette changes, while human evaluations provide complementary evidence of perceptual quality. Our main contributions are summarized as follows:

\begin{itemize}

\item We formulate source-garment interference as a conditioning problem and introduce ToWo, a two-stage Take-Off/Wear-On paradigm that separates body-proxy generation from target-garment synthesis.

\item We develop the Body Structure Revealing (BSR) model and construct the Body-Fitting DressCode (BFDC) dataset, aligning body-proxy conditioning between inference and training.

\item We adapt Wear-On to both diffusion-based VTON models and an MLLM-guided DiT by conditioning garment synthesis on the body proxy. In the MLLM variant, GBA supplies fine-grained garment features to the denoiser. 

\item We validate ToWo-VTON across upper-body, lower-body, and full-body try-on, reporting image-quality improvements over the evaluated baselines and complementary qualitative and human evaluations.
\end{itemize}

\section{Related Works}
\label{sec:rel}

\noindent\textbf{GAN-based VTON.}
Early VTON methods relied on GANs \citep{goodfellow2020generative} with appearance flow and garment warping \citep{Xie_2023_CVPR, He_2022_CVPR, Chopra_2021_ICCV, Yu_2019_ICCV, Dong_2019_ICCV, Ge_2021_CVPR, Chen_2021_ICCV, Wang_2018_ECCV}. VITON \citep{Han_2018_CVPR} introduced pose-guided content warping for garment alignment, followed by condition refinement to better estimate body–garment correspondence \citep{Cui_2021_ICCV, Yang_2020_CVPR}. Later works like SD-VITON \citep{10.1609/aaai.v38i5.28288} incorporated pose maps for improved alignment. S-WUTON \citep{10.1007/978-3-030-58565-5_37} introduced a teacher–student paradigm for garment transfer. However, GAN-based methods often suffer from limited generalization and visual artifacts.

\vspace{0.5em}
\noindent\textbf{Diffusion-based VTON.}
Diffusion models have recently enabled more flexible and realistic garment transfer through denoising-based generation \citep{Yang_2024_CVPR, Liang_2025_CVPR, Hong_2025_CVPR}. Unlike warping-based methods, they leverage image inpainting to integrate garments more naturally \citep{wan_incorporating_2025, Yan_2023_CVPR, Li_2023_ICCV, Kim_2025_ICCV, Zhang_2025_ICCV}. Works such as StableVITON \citep{Kim_2024_CVPR} and IDM-VTON \citep{10.1007/978-3-031-73016-0_13} improve identity preservation via structured conditioning, but often require additional inputs (e.g., pose maps or 3D cues \citep{He_2025_CVPR}) that introduce computational overhead and noise sensitivity. To reduce reliance on manual inputs, recent methods adopt data-driven pipelines with enhanced identity and text control \citep{chong_catvton_2025, Du_2025_ICCV, Zhang_2025_CVPR, Wan_2025_ICCV, ye2023ipadaptertextcompatibleimage, wan2026mobile}. Yet, most works focus on upper-body garments, with lower-body VTON remaining challenging due to garment diversity (e.g., skirts, dresses) and occlusion handling \citep{Chen_2024_ECCV, Yang_2025_ICCV, Zhu_2024_CVPR, Li_2025_CVPR, Guo_2025_ICCV, Chen_2023_ICCV}.

\vspace{0.5em}
\noindent\textbf{MLLMs for Image Editing.}
MLLMs support vision-language tasks, including VTON, by aligning visual and textual features \citep{wu2025omnigen2explorationadvancedmultimodal}. Models like GenArtist \citep{NEURIPS2024_e7c78602} and VITRON \citep{NEURIPS2024_68bad550} enable fine-grained control via language prompts, while Qwen-Image \citep{wu2025qwenimagetechnicalreport} and DreamOmni2 \citep{xia2025dreamomni2multimodalinstructionbasedediting} offer cross-modal representations. Our MLLM-guided implementation adapts OmniGen2's text and reference-image conditioning to the body-proxy try-on task. Its native alignment components are inherited from the backbone; our additions are the ToWo conditioning strategy and GBA for garment details.

\section{Methodology}

\subsection{Body Topology Generation via Take-Off}
\label{sec:towear}

Source-garment contours and coverage can bias target-garment synthesis even when the person's overall pose is visible. Take-Off reduces this interference by generating a tight-fitting clothed proxy before Wear-On synthesis. The proxy represents visible body contours, limb layout, and pose in image space; it is an estimated structural condition, not ground-truth hidden anatomy. Its image representation can be passed to existing person-conditioning branches without introducing a separate pose or mesh interface.

\begin{figure*}[tb]
    \centering
    \includegraphics[width=1.0\textwidth]{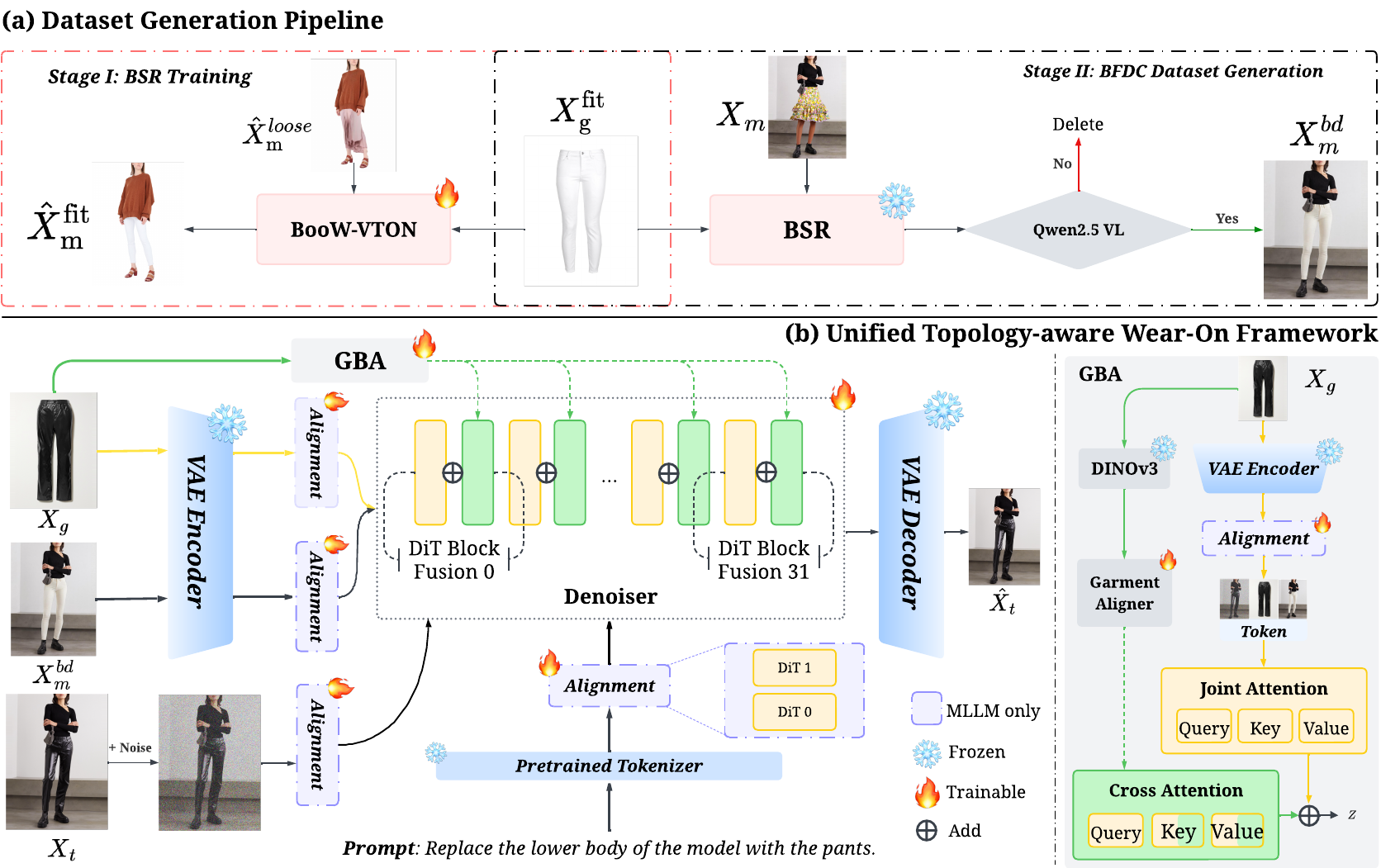}
    \caption{\textbf{Overview of the ToWo-VTON framework.} (a) \textit{Dataset Generation}: Stage I uses reproduced BooW-VTON to create loose-clothing variants from tight-fit samples, enabling fine-tuning into the Body Structure Revealing (BSR) model. Stage II applies BSR to generate tight-fit surrogates for DressCode images, forming the BFDC dataset after filtering by Qwen2.5-VL. (b) \textit{Unified Topology-aware Wear-On framework}: Wear-On uses the Take-Off surrogate $X_m^{\text{bd}}$ as the body-conditioning input for garment synthesis. Existing diffusion-based VTON backbones can be adapted by replacing the garment-entangled person image $X_m$ with $X_m^{\text{bd}}$, while the MLLM variant uses the proxy and garment as reference images. The boxes labeled Alignment denote OmniGen2's native input refiners.}
    \label{fig:model}
    \vspace{-1.0em}
\end{figure*}

\vspace{0.5em}
\noindent\textbf{BSR model.}
Take-Off is operationalized through the Body Structure Revealing (BSR) model.
Given a person image with potentially loose or occluding clothing, BSR generates a tight-fitting surrogate intended to retain identity and pose while reducing source-garment coverage.
This surrogate is not intended to reconstruct private body appearance, but rather to expose the geometric cues required for topology-aware garment synthesis.

To obtain BSR, we reproduce BooW-VTON~\citep{Zhang_2025_CVPRBOOW} and fine-tune it for tight-fit body surrogate reconstruction.
Specifically, we construct $\mathcal{D}_{\text{BF}}$ exclusively from the official DressCode training split~\citep{10.1007/978-3-031-20074-8_20}. After VLM-based screening for tight-fitting clothing, we retain the 3,000 highest-confidence person--garment pairs from each of the upper-body and lower-body categories, yielding 6,000 training pairs $(X_m^{\text{fit}}, X_g^{\text{fit}})$ in total.
For each category, we also manually select ten tight-fitting garment templates, giving twenty templates in total. The templates are visualized in the supplementary material.

As illustrated in Fig.~\ref{fig:model}(a) Stage I, we synthesize occluded inputs by re-dressing each tight-fitting person image $X_m^{\text{fit}}$ with a randomly sampled garment $X_g$ and its textual description $D_g$ using our reproduced BooW-VTON.
This produces a loose-clothing variant $\hat{X}_m^{\text{loose}}$ that simulates realistic garment-induced occlusion while preserving the underlying subject identity and pose.
We then form training triplets $(X_g^{\text{fit}}, X_m^{\text{fit}}, \hat{X}_m^{\text{loose}})$, where $X_g^{\text{fit}}$ serves as the tight-fit template condition, $X_m^{\text{fit}}$ is the desired reconstruction target, and $\hat{X}_m^{\text{loose}}$ is the occluded input to be restored.

Using these triplets, we fine-tune the reproduced BooW-VTON into the BSR model, which reconstructs the tight-fitting reference from the occluded input conditioned on the tight-fit template and its associated textual description:
\begin{equation}
\hat{X}_m^{\text{fit}} = \text{BSR}(\hat{X}_m^{\text{loose}} \mid X_g^{\text{fit}}, D_g^{\text{fit}}).
\end{equation}
The model is optimized using the original diffusion objective, encouraging BSR to suppress garment-induced ambiguity while preserving identity, pose, and body structure.

\begin{figure*}[t]
     \centering
     \includegraphics[width=0.9\textwidth]{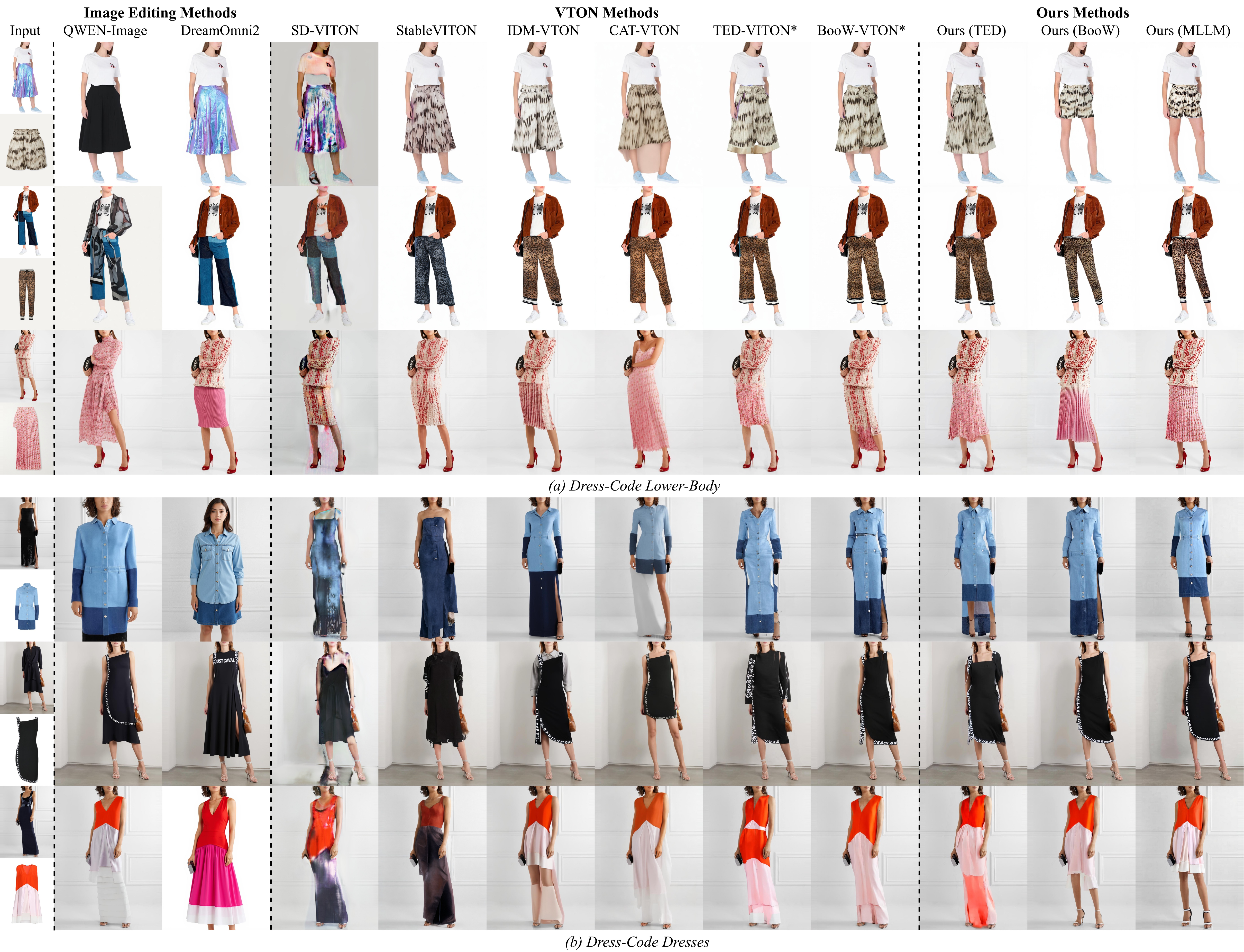}
     \caption{Qualitative comparison on (a) DressCode Lower-body and (b) DressCode Dresses. Ours (TED) and Ours (BooW) are topology-aware diffusion variants, while Ours is the MLLM-guided ToWo-VTON. Across garment categories, our models exhibit more consistent garment–body alignment and clearer garment details than competing approaches. The improvements are reflected in more coherent garment placement and preservation of fine-grained characteristics, yielding try-on results that maintain structural consistency and garment fidelity.}
     \label{fig:comp}
     \vspace{-1.0em}
\end{figure*}

\vspace{0.5em}
\noindent\textbf{Body-Fitting DressCode (BFDC) Dataset.}
While Take-Off provides a body proxy decoupled from the source garment at inference, Wear-On synthesis requires the generator to be trained with the same type of structural conditioning.
If the generator is trained only on garment-entangled person images but tested with Take-Off surrogates, the distribution gap may weaken the benefit of topology-aware conditioning.
Therefore, we construct Body-Fitting DressCode (BFDC), a geometry-aware training set that extends structural decoupling from inference to training.

As illustrated in Fig.~\ref{fig:model}(a) Stage II, BFDC is constructed strictly from the official DressCode training split~\citep{10.1007/978-3-031-20074-8_20}. The 1,800 test pairs in each category are excluded before generating training proxies. For each training person image $X_m$, we apply the trained BSR model using a selected tight-fit template $(X_g^{\text{fit}}, D_g^{\text{fit}})$:
\begin{equation}
X_m^{\text{bd}} = \text{BSR}(X_m \mid X_g^{\text{fit}}, D_g^{\text{fit}}).
\label{eq:bfdc_hat}
\end{equation}
For dress images, we perform upper-body and lower-body replacements in successive BSR applications using the corresponding tight-fit templates, passing the output of one step to the next. The final output serves as $X_m^{\text{bd}}$.
The resulting surrogate, generated at the original DressCode resolution, is intended to retain identity and pose while replacing the source garment with tight-fitting clothing. We form training triplets $(X_m, X_g, X_m^{\text{bd}})$, where $X_m^{\text{bd}}$ provides geometry-aware conditioning for Wear-On.

We filter the generated proxies using Qwen2.5-VL~\citep{qwen_qwen25_2025}, guided by valid and invalid exemplars with rationales. The filter checks for insufficiently tight clothing, disconnected or duplicated limbs, and overlapping clothing artifacts, including whether lower-body proxies expose leg contours. Only proxies receiving a positive binary decision are retained; this screening assesses visible plausibility rather than anatomical accuracy.
The retained dataset, denoted as $\mathcal{D}_{\text{BFDC}}$, supplies body-proxy conditioning paired with original-image targets for upper-body, lower-body, and full-body try-on.
Take-Off and $\mathcal{D}_{\text{BFDC}}$ establish a unified structural decoupling paradigm: Take-Off supplies body proxies at inference time, while $\mathcal{D}_{\text{BFDC}}$ aligns the Wear-On generator with the same type of conditioning during training.

\subsection{Unified Topology-aware Wear-On Framework}

In the Wear-On stage, we use $X_m^{\text{bd}}$ to denote the body proxy produced by Take-Off, either from $\mathcal{D}_{\text{BFDC}}$ during training or from the input person image during inference. 
Wear-On synthesizes the target try-on image using $X_m^{\text{bd}}$ as structural conditioning. 
Compared with the person image $X_m$, which entangles body structure with source-garment appearance, $X_m^{\text{bd}}$ provides a cleaner topology carrier, including pose, body contour, limb configuration, and layout. 
By replacing $X_m$ with $X_m^{\text{bd}}$, Wear-On passes the estimated body structure into garment synthesis, while $X_g$ and the garment-related prompt provide appearance and semantic guidance.

Given a garment-related prompt $P$, a frozen pretrained encoder produces semantic tokens:
\begin{equation}
T = \mathcal{P}_{\text{enc}}(P),
\end{equation}
where $\mathcal{P}_{\text{enc}}$ denotes the pretrained encoder used for prompt encoding.
For diffusion-based VTON backbones, $\mathcal{P}_{\text{enc}}$ follows the original CLIP/T5 encoders.
For our MLLM-guided design, it is instantiated with Qwen2.5-VL to provide stronger vision-grounded semantic guidance.

At an abstract level, Wear-On performs conditional denoising with three complementary sources of information: semantic guidance from $T$, body topology from $X_m^{\text{bd}}$, and garment appearance from $X_g$. 
This generic conditioning interface can be written as:
\begin{equation}
\hat{\epsilon} = 
\mathcal{D}_{\theta}
\left(
\mathbf{z}_t, t \mid T, \mathcal{E}(X_m^{\text{bd}}), \mathcal{E}(X_g)
\right),
\end{equation}
where $\mathcal{D}_{\theta}$ denotes the diffusion denoiser and $\mathcal{E}$ denotes a pretrained VAE encoder. 
This conditioning interface applies to both the diffusion-based VTON backbones and the MLLM-guided DiT in Fig.~\ref{fig:model}(b).

\begin{table*}[t]
\centering
\footnotesize
\setlength{\tabcolsep}{4pt}
\renewcommand{\arraystretch}{1.18}

\caption{Quantitative results on the DressCode Lower-body and Dresses datasets. Ours (TED) and Ours (BooW) are topology-aware plug-in variants, while Ours is the MLLM-guided variant. w/o ToWo retains GBA and uses original person images during training and inference. \textbf{Bold} and \underline{underline} indicate the best and second-best results. FID is scaled by 100. * indicates reproduced results.}
\label{tab:cross_metrics}

\begin{tabularx}{\textwidth}{l*{8}{>{\centering\arraybackslash}X}}
\specialrule{0.8pt}{0pt}{0pt}
\textbf{Dataset}
& \multicolumn{4}{c}{\textbf{Dress-Code Lower-body}}
& \multicolumn{4}{c}{\textbf{Dress-Code Dresses}} \\
\cmidrule(lr){2-5}\cmidrule(lr){6-9}
\textbf{Method}
& KID$\downarrow$ & FID$\downarrow$
& SSIM$\uparrow$ & LPIPS$\downarrow$
& KID$\downarrow$ & FID$\downarrow$
& SSIM$\uparrow$ & LPIPS$\downarrow$ \\

\hline
\rowcolor[HTML]{EFEFEF}
\multicolumn{9}{c}{\textbf{Image Editing Methods}} \\
\hline
QWEN-Image \citep{wu2025qwenimagetechnicalreport}
& 0.907 & 19.438 & 0.842 & 0.212
& 0.389 & 15.667 & 0.826 & 0.189 \\
DreamOmni2 \citep{xia2025dreamomni2multimodalinstructionbasedediting}
& 0.848 & 21.669 & 0.801 & 0.286
& 1.900 & 35.636 & 0.738 & 0.301 \\

\hline
\rowcolor[HTML]{EFEFEF}
\multicolumn{9}{c}{\textbf{GAN-based VTON Methods}} \\
\hline
SD-VITON \citep{10.1609/aaai.v38i5.28288}
& 6.202 & 69.156 & 0.825 & 0.370
& 8.830 & 85.921 & 0.801 & 0.378 \\

\hline
\rowcolor[HTML]{EFEFEF}
\multicolumn{9}{c}{\textbf{Diffusion-based VTON Methods}} \\
\hline
StableVITON \citep{Kim_2024_CVPR}
& 0.966 & 19.732 & 0.878 & 0.216
& 0.707 & 23.454 & 0.850 & 0.210 \\
IDM-VTON \citep{10.1007/978-3-031-73016-0_13}
& 0.497 & 16.125 & 0.871 & 0.193
& \cellcolor[HTML]{FFCE93}\underline{0.359}
& \cellcolor[HTML]{FFCE93}\underline{14.065}
& \cellcolor[HTML]{FFCE93}\underline{0.860}
& 0.173 \\
CAT-VTON \citep{chong_catvton_2025}
& 1.012 & 20.146 & 0.875 & 0.232
& 0.640 & 17.210 & 0.841 & 0.211 \\
TED-VITON* \citep{wan2025tedvitontransformerempowereddiffusionmodels}
& 0.762 & 19.500 & 0.877 & 0.184
& 0.497 & 15.032 & 0.846 & 0.182 \\
BooW-VTON* \citep{Zhang_2025_CVPRBOOW}
& 0.509 & 16.724 & 0.854 & 0.210
& 0.583 & 17.526 & 0.844 & 0.172 \\
\textbf{Ours (TED)}
& 0.584 & 14.663 & 0.890 & 0.183
& 0.408 & 14.926
& \cellcolor[HTML]{FFFC9E}\textbf{0.865}
& 0.168 \\
\textbf{Ours (BooW)}
& \cellcolor[HTML]{FFCE93}\underline{0.376}
& \cellcolor[HTML]{FFCE93}\underline{13.952}
& \cellcolor[HTML]{FFCE93}\underline{0.892}
& \cellcolor[HTML]{FFCE93}\underline{0.172}
& 0.491 & 15.849 & 0.858
& \cellcolor[HTML]{FFCE93}\underline{0.166} \\

\hline
\rowcolor[HTML]{EFEFEF}
\multicolumn{9}{c}{\textbf{MLLM-guided VTON Methods}} \\
\hline
Ours w/o ToWo
& 0.415 & 14.201 & 0.877 & 0.179
& 1.026 & 23.007 & 0.826 & 0.200 \\
\textbf{Ours}
& \cellcolor[HTML]{FFFC9E}\textbf{0.172}
& \cellcolor[HTML]{FFFC9E}\textbf{11.044}
& \cellcolor[HTML]{FFFC9E}\textbf{0.895}
& \cellcolor[HTML]{FFFC9E}\textbf{0.154}
& \cellcolor[HTML]{FFFC9E}\textbf{0.190}
& \cellcolor[HTML]{FFFC9E}\textbf{12.459}
& \cellcolor[HTML]{FFCE93}\underline{0.860}
& \cellcolor[HTML]{FFFC9E}\textbf{0.152} \\
\specialrule{0.8pt}{0pt}{0pt}
\end{tabularx}
\end{table*}

\subsubsection{Plug-in Adaptation to Diffusion-based Backbones}

We instantiate this plug-in design on two representative diffusion-based VTON backbones: the mask-based TED-VITON~\citep{wan2025tedvitontransformerempowereddiffusionmodels} and the mask-free BooW-VTON~\citep{Zhang_2025_CVPRBOOW}. 
Both backbones already contain a person-conditioning branch, allowing the Take-Off body proxy to be used as the structural input without changing the network architecture, diffusion objective, or sampling pipeline.
Given a diffusion-based VTON generator $\mathcal{F}$, the adapted process is:
\begin{equation}
\hat{X}_t = \mathcal{F}(X_m^{\text{bd}} \mid X_g, D_g),
\label{eq:plug}
\end{equation}
where $X_g$ provides target garment appearance, $D_g$ provides semantic guidance, and $X_m^{\text{bd}}$ provides decoupled body topology.
We further fine-tune each backbone on $\mathcal{D}_{\text{BFDC}}$ to align the training distribution with the Take-Off conditioning used at inference time. 
This enables the generator to interpret $X_m^{\text{bd}}$ as its structural input while reducing reliance on source-garment silhouettes.

\subsubsection{Scaling to MLLM-guided Backbones}

We adapt OmniGen2~\citep{wu2025omnigen2explorationadvancedmultimodal} to Wear-On by supplying the Take-Off body proxy $X_m^{\text{bd}}$ and the target garment $X_g$ as its two reference images. The proxy provides the body geometry, and the garment provides the target appearance.

\vspace{0.5em}
\noindent\textbf{Backbone input conditioning.}
OmniGen2's native input refiners prepare the inputs to its main DiT at each denoising step. The context refiner processes the Qwen2.5-VL prompt tokens $T$, the reference-image refiner processes the two image latents $\mathcal{E}(X_m^{\text{bd}})$ and $\mathcal{E}(X_g)$ separately, and the noise refiner processes the current noisy latent $\mathbf{z}_t$. The backbone then concatenates the resulting tokens for joint processing in the main DiT. These refiners (the alignment blocks in Fig.~\ref{fig:model}) are inherited from OmniGen2, with no additional alignment-module design claimed here. All variants retain them; deleting them would disrupt the backbone's input pathway rather than isolate a ToWo contribution.

\vspace{0.5em}
\noindent\textbf{Topology-conditioned Denoiser.}
At each step, the main DiT jointly processes the refined tokens to predict the denoising update. The body-proxy tokens carry the estimated pose and body contour, while the garment and prompt tokens specify what the person should wear.

\vspace{0.5em}
\noindent\textbf{Garment Boosting Attention (GBA).}
To preserve garment details that may be weakened during denoising, such as texture and trim, we introduce Garment Boosting Attention (GBA). GBA extracts garment features with DINOv3~\citep{siméoni2025dinov3} and maps them to garment-specific keys and values using a learnable Garment Aligner $G$,
\begin{equation}
(K_g, V_g) = G(\text{DINOv3}(X_g)).
\end{equation}
Each DiT layer integrates these features via cross-attention:
\begin{equation}
Z = \text{Attention}(Q, K_j, V_j) + \lambda \, \text{Attention}(Q, K_g, V_g),
\label{eq:attn}
\end{equation}
where $Q$, $K_j$, and $V_j$ are from the main denoising stream, and $\lambda$ controls the strength of garment feature fusion. 
By injecting DINOv3 garment features into the topology-conditioned denoising stream, GBA provides each DiT layer with fine-grained garment cues, enhancing texture fidelity, preserving local details, and improving garment-body coherence under body-proxy conditioning.

\section{Experiments}

\subsection{Experiment Setup}

\noindent\textbf{Datasets.}
We evaluate our model on two VTON benchmarks: VITON-HD~\citep{Choi_2021_CVPR} and DressCode~\citep{10.1007/978-3-031-20074-8_20}. We use reproduced BooW-VTON~\citep{Zhang_2025_CVPRBOOW} to construct re-dressed inputs for mask-free paired reconstruction. Input construction differs between paired and unpaired evaluation as specified below. We adopt the official training and test splits. For Dress-Code, 1,800 pairs are selected for testing from each subset: 8,951 lower-body, 29,478 dress, and 15,363 upper-body pairs. For VITON-HD, 2,032 of 13,679 pairs are used for testing, with the remaining samples utilized for training. All images are processed at a resolution of 768 $\times$ 1024 to ensure consistency across models and datasets.

\begin{table*}[t]
\centering
\footnotesize
\setlength{\tabcolsep}{4pt}
\renewcommand{\arraystretch}{1.18}

\caption{Quantitative comparison on VITON-HD and DressCode Upper-body benchmarks. Ours (TED) and Ours (BooW) are topology-aware plug-in variants, while Ours is the MLLM-guided variant. w/o ToWo retains GBA and uses original person images during training and inference. \textbf{Bold} and \underline{underline} indicate the best and second-best results. FID is scaled by 100. * indicates reproduced results.}
\label{tab:VITON_HD}

\begin{tabularx}{\textwidth}{l*{8}{>{\centering\arraybackslash}X}}
\specialrule{0.8pt}{0pt}{0pt}
\textbf{Dataset}
& \multicolumn{4}{c}{\textbf{VITON-HD}}
& \multicolumn{4}{c}{\textbf{Dress-Code Upper-body}} \\
\cmidrule(lr){2-5}\cmidrule(lr){6-9}
\textbf{Method}
& KID$\downarrow$ & FID$\downarrow$
& SSIM$\uparrow$ & LPIPS$\downarrow$
& KID$\downarrow$ & FID$\downarrow$
& SSIM$\uparrow$ & LPIPS$\downarrow$ \\

\hline
\rowcolor[HTML]{EFEFEF}
\multicolumn{9}{c}{\textbf{Image Editing Methods}} \\
\hline
QWEN-Image \citep{wu2025qwenimagetechnicalreport}
& 0.794 & 16.499 & 0.688 & 0.381
& 2.311 & 32.957 & 0.732 & 0.372 \\
DreamOmni2 \citep{xia2025dreamomni2multimodalinstructionbasedediting}
& 0.551 & 15.076 & 0.743 & 0.345
& 1.528 & 30.054 & 0.771 & 0.325 \\

\hline
\rowcolor[HTML]{EFEFEF}
\multicolumn{9}{c}{\textbf{GAN-based VTON Methods}} \\
\hline
SD-VITON \citep{10.1609/aaai.v38i5.28288}
& 0.280 & 10.552
& \cellcolor[HTML]{FFCE93}\underline{0.864}
& 0.204
& 6.143 & 67.833 & 0.847 & 0.335 \\

\hline
\rowcolor[HTML]{EFEFEF}
\multicolumn{9}{c}{\textbf{Diffusion-based VTON Methods}} \\
\hline
StableVITON \citep{Kim_2024_CVPR}
& 0.575 & 14.105 & 0.843 & 0.232
& 0.920 & 19.112 & 0.890 & 0.222 \\
IDM-VTON \citep{10.1007/978-3-031-73016-0_13}
& 0.423 & 12.248 & 0.839 & 0.225
& 0.304 & 11.821 & 0.915 & 0.160 \\
CAT-VTON \citep{chong_catvton_2025}
& 0.554 & 12.847 & 0.828 & 0.266
& 0.645 & 14.550 & 0.897 & 0.208 \\
TED-VITON* \citep{wan2025tedvitontransformerempowereddiffusionmodels}
& 0.359 & 12.063 & 0.845 & 0.221
& 0.373 & 13.480 & 0.909 & 0.178 \\
BooW-VTON* \citep{Zhang_2025_CVPRBOOW}
& 0.254 & 11.621 & 0.830 & 0.234
& 0.289 & 10.642 & 0.915 & 0.170 \\
\textbf{Ours (TED)}
& \cellcolor[HTML]{FFCE93}\underline{0.184}
& \cellcolor[HTML]{FFCE93}\underline{8.968}
& 0.863
& \cellcolor[HTML]{FFCE93}\underline{0.200}
& 0.278 & 12.107
& \cellcolor[HTML]{FFFC9E}\textbf{0.922}
& 0.186 \\
\textbf{Ours (BooW)}
& 0.194 & 9.868 & 0.847 & 0.231
& \cellcolor[HTML]{FFCE93}\underline{0.220}
& \cellcolor[HTML]{FFFC9E}\textbf{9.815}
& 0.918
& \cellcolor[HTML]{FFCE93}\underline{0.157} \\

\hline
\rowcolor[HTML]{EFEFEF}
\multicolumn{9}{c}{\textbf{MLLM-guided VTON Methods}} \\
\hline
Ours w/o ToWo
& 0.225 & 10.028 & 0.826 & 0.246
& 0.296 & 13.834 & 0.902 & 0.189 \\
\textbf{Ours}
& \cellcolor[HTML]{FFFC9E}\textbf{0.176}
& \cellcolor[HTML]{FFFC9E}\textbf{8.673}
& \cellcolor[HTML]{FFFC9E}\textbf{0.868}
& \cellcolor[HTML]{FFFC9E}\textbf{0.195}
& \cellcolor[HTML]{FFFC9E}\textbf{0.194}
& \cellcolor[HTML]{FFCE93}\underline{10.532}
& \cellcolor[HTML]{FFCE93}\underline{0.920}
& \cellcolor[HTML]{FFFC9E}\textbf{0.149} \\
\specialrule{0.8pt}{0pt}{0pt}
\end{tabularx}
\vspace{-1.0em}
\end{table*}

\vspace{0.5em}
\noindent\textbf{Baselines.}
We compare ToWo-VTON with representative methods across four categories. 
For image editing approaches, we include recent multimodal generative systems QWEN-Image~\citep{wu2025qwenimagetechnicalreport} and DreamOmni2~\citep{xia2025dreamomni2multimodalinstructionbasedediting}. 
For GAN-based VTON, we adopt SD-VITON~\citep{10.1609/aaai.v38i5.28288}. 
Diffusion-based VTON baselines include StableVITON~\citep{Kim_2024_CVPR}, IDM-VTON~\citep{10.1007/978-3-031-73016-0_13}, CAT-VTON~\citep{chong_catvton_2025}, TED-VITON~\citep{wan2025tedvitontransformerempowereddiffusionmodels}, and BooW-VTON~\citep{Zhang_2025_CVPRBOOW}. 
For fair comparison, we reproduce TED-VITON and BooW-VTON under a unified evaluation protocol. 
To evaluate the plug-in capability of our topology-aware framework, we apply Take-Off to two representative diffusion paradigms and fine-tune them on $\mathcal{D}_{\text{BFDC}}$, namely the mask-based TED-VITON and the mask-free BooW-VTON, denoted as Ours (TED) and Ours (BooW), respectively. 
These variants retain the original architectures, training objectives, and diffusion schedules, differing only in the proposed topology-aware conditioning.

We adapt OmniGen2 to direct Wear-On, denoted as Ours w/o ToWo, retaining GBA and using original person images during training and inference. The full model replaces these inputs with Take-Off body proxies under matched training settings, as detailed in Section~\ref{sec:ablation_study}.
All models are evaluated at a unified resolution of $768 \times 1024$, with outputs rescaled when necessary to ensure fairness. 
A detailed analysis of the computational overhead introduced by the Take-Off stage and the MLLM-guided Wear-On implementation is provided in Supplementary.

\vspace{0.5em}
\noindent\textbf{Evaluation Metrics.}
We use Fréchet Inception Distance (FID)~\citep{heusel2018ganstrainedtimescaleupdate} and Kernel Inception Distance (KID)~\citep{bińkowski2021demystifyingmmdgans} for distributional image quality, and Structural Similarity Index (SSIM)~\citep{1284395} and Learned Perceptual Image Patch Similarity (LPIPS)~\citep{zhang2018unreasonableeffectivenessdeepfeatures} for reference-image similarity. KID/FID evaluate unpaired generation, while SSIM/LPIPS evaluate paired reconstruction. In paired evaluation, mask-based methods reconstruct the masked garment region; mask-free methods first receive a synthetically re-dressed person and then regenerate the full original image. These interfaces therefore impose different reconstruction demands. In unpaired evaluation, both use a target garment different from the input clothing. Reported FID values equal the raw scores multiplied by 100. These image-quality metrics do not directly measure limb connectivity, pose error, or anatomical accuracy. Qualitative comparisons and the supplementary user study provide complementary evidence about visible garment fitting and body layout.

\begin{figure}[t]
    \centering
    \includegraphics[width=\linewidth]{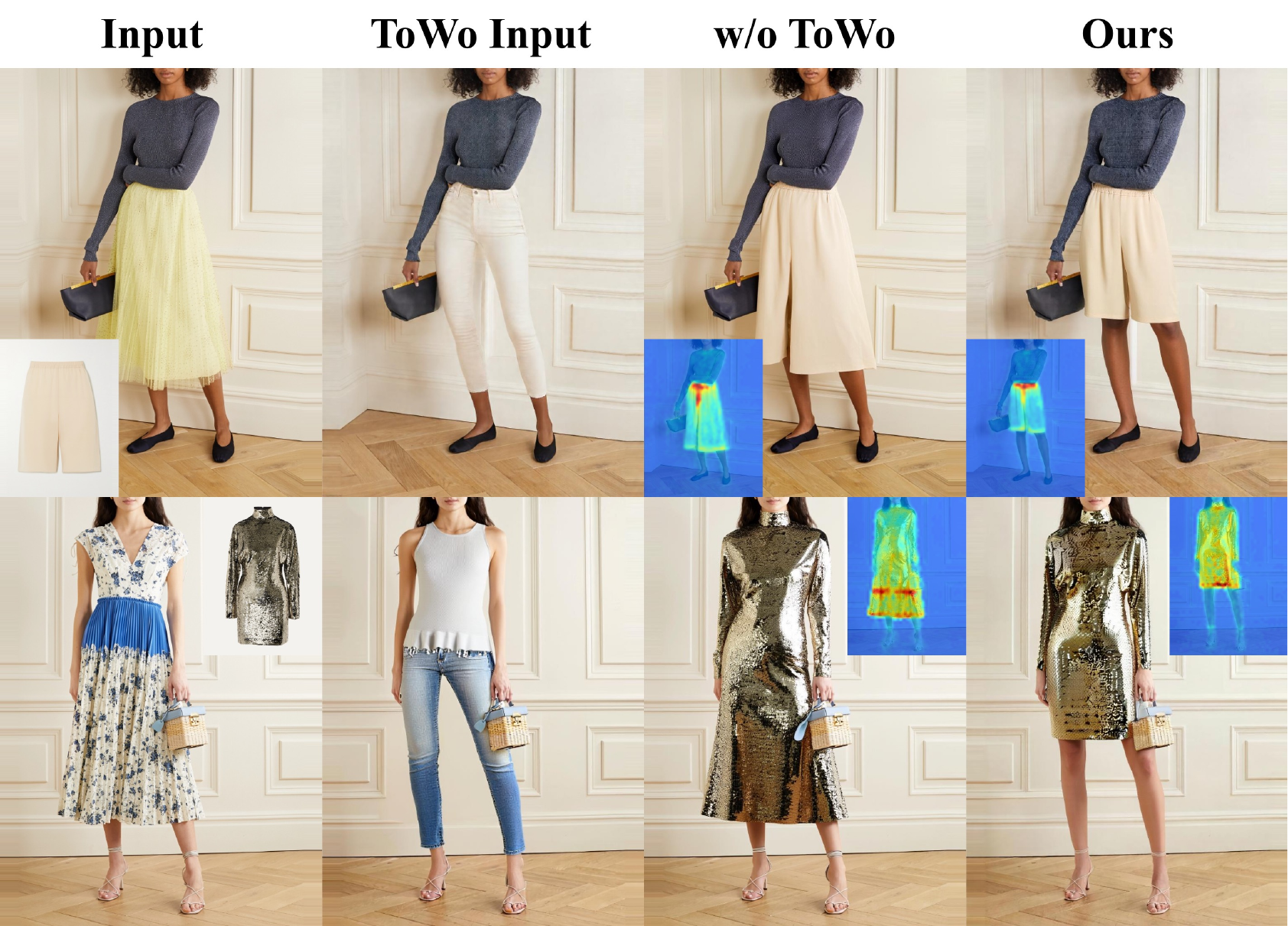}
    \caption{\textbf{Topology-aware attention analysis.}
    ToWo Input shows the body proxy generated by Take-Off for Wear-On.
    Attention shifts toward body-structure cues with ToWo.
    \textit{w/o ToWo} retains GBA, using original person images during training and inference.}
    \label{fig:attention_map}
    \vspace{-1.0em}
\end{figure}

\subsection{Qualitative Results}

Figure~\ref{fig:comp} presents qualitative comparisons on lower-body and dress cases involving changes in garment length, looseness, and coverage. In lower-body try-on, competing methods sometimes retain source-clothing boundaries or produce distorted leg regions, particularly when the target garment exposes areas previously covered by the source garment. ToWo results show cleaner garment boundaries and more coherent visible limb layouts, supporting a more natural transition between clothing and exposed body regions. In dress cases, changes in garment length and silhouette affect a larger portion of the body, requiring consistent garment placement across these regions. The illustrated ToWo results exhibit more coherent garment outlines and body--garment relationships under these transformations. Together, these examples suggest that conditioning synthesis on the body proxy helps reduce source-garment interference and accommodate the target garment's coverage. Additional upper-body and background-varied examples are provided in the supplementary material. Section~\ref{sec:ablation_study} further examines the combined ToWo strategy and the contribution of GBA to garment-detail preservation.

\subsection{Quantitative Results}

Tables~\ref{tab:cross_metrics} and~\ref{tab:VITON_HD} report quantitative image-quality comparisons under our VITON-HD and DressCode evaluation protocols. The full ToWo-VTON model achieves the best results on all four metrics on DressCode Lower-body. On DressCode Dresses, it leads KID, FID, and LPIPS while tying for second-best SSIM, demonstrating competitive performance across both distribution-level and reconstruction metrics. For diffusion-based implementations, Ours (TED) and Ours (BooW) consistently improve KID and FID over their corresponding reimplementations in every reported category, supporting the applicability of body-proxy conditioning to both mask-based and mask-free backbones. The full MLLM-guided model also improves all four metrics over Ours w/o ToWo across the reported benchmarks. Together, these comparisons support the combined BFDC-training and two-stage inference strategy across the evaluated architectures and garment categories. The scope of this joint comparison is detailed below.

\begin{figure}[t]
    \centering
    \includegraphics[width=\linewidth]{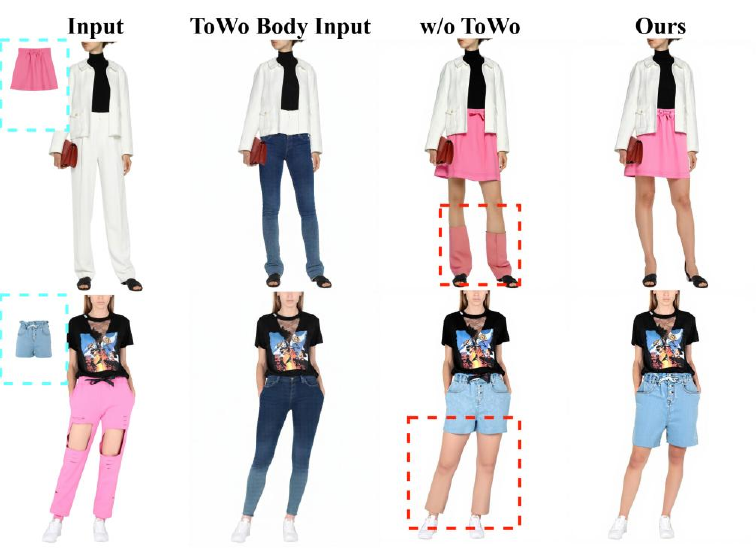}
    \caption{\textbf{Effect of ToWo.}
    The first column shows the original person image and target garment.
    The second column, labeled \textit{ToWo Body Input}, shows the body proxy generated by the Take-Off stage, which serves as the structural conditioning input to the subsequent Wear-On stage.}
    \label{fig:ablation_towo}
    \vspace{-1.0em}
\end{figure}

\begin{figure}[t]
    \centering
    \includegraphics[width=\linewidth]{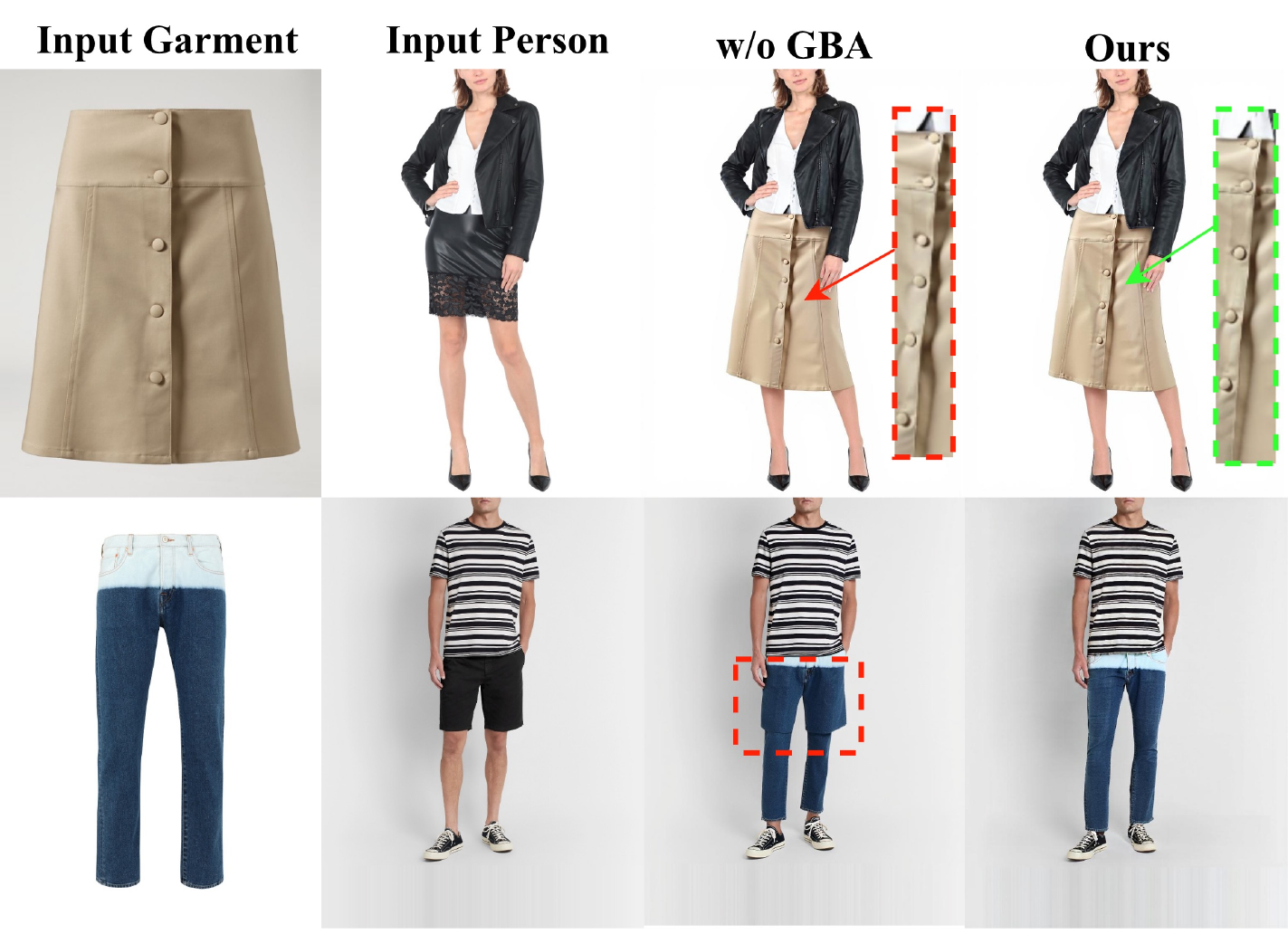}
    \caption{\textbf{Effect of GBA.}
    The first two columns show the input garment and person.
    Zoom in for finer details.}
    \label{fig:ablation_gba}
    \vspace{-1.0em}
\end{figure}

\subsection{Topology Awareness Analysis}

Figure~\ref{fig:attention_map} visualizes attention patterns under direct person-image conditioning and the ToWo strategy. In the direct baseline, attention concentrates on source-clothing boundaries, suggesting that the garment silhouette remains a conditioning cue. With ToWo, attention is distributed around the waist, hips, and leg contours, which provide spatial cues for target garment placement and body--garment alignment. This shift is consistent with the intended role of the tight-fitting proxy in making body structure more accessible while reducing source-garment interference. Together with the qualitative results, these visualizations help characterize how the model responds to body-proxy conditioning. Since the compared variants differ in both training and inference conditioning, the observed patterns provide descriptive evidence for the combined strategy.

\begin{table}[t]
    \centering
    \footnotesize
    \setlength{\tabcolsep}{4pt}
    \renewcommand{\arraystretch}{1.2}

    \caption{Ablation study on ToWo and GBA.
    \textbf{Bold} indicates the best result.
    FID is scaled by 100.}
    \label{tab:ablation}

    \begin{tabularx}{\linewidth}{
        @{}l*{4}{>{\centering\arraybackslash}X}@{}
    }
        \toprule
        \textbf{Method}
        & KID$\downarrow$
        & FID$\downarrow$
        & SSIM$\uparrow$
        & LPIPS$\downarrow$ \\
        \midrule
        w/o ToWo
        & 0.415 & 14.201 & 0.877 & 0.179 \\
        w/o GBA
        & 0.586 & 17.775 & 0.857 & 0.201 \\
        \midrule
        \textbf{Ours}
        & \textbf{0.172}
        & \textbf{11.044}
        & \textbf{0.895}
        & \textbf{0.154} \\
        \bottomrule
    \end{tabularx}
    \vspace{-2.0em}
\end{table}

\subsection{Ablation Studies}
\label{sec:ablation_study}


We evaluate ToWo and GBA on DressCode Lower-body. All variants retain OmniGen2's native input refiners as inherited backbone components. The w/o ToWo variant retains GBA but uses original person images instead of body proxies during training and performs direct Wear-On inference without Take-Off. This variant and the full model share the same retained person--garment pairs, initialization, trainable modules, optimization settings, and number of training updates. Their comparison assesses body-proxy conditioning across training and inference jointly, without isolating the two effects. Table~\ref{tab:ablation} shows improvements in all four metrics with ToWo, while Figure~\ref{fig:ablation_towo} illustrates cleaner limb and garment boundaries. Corresponding diffusion-backbone comparisons appear in Tables~\ref{tab:cross_metrics} and~\ref{tab:VITON_HD}. The w/o GBA variant retains BFDC training and Take-Off/Wear-On inference, removing only the added garment-feature branch. Within this configuration, adding GBA reduces LPIPS from 0.201 to 0.154 and increases SSIM from 0.857 to 0.895. Figure~\ref{fig:ablation_gba} illustrates improvements in buttons and pant outlines, supporting GBA's role in garment-detail conditioning.

\section{Conclusions}
We presented ToWo-VTON, a two-stage framework that reduces source-garment interference through body-proxy conditioning. Take-Off generates a tight-fitting proxy for Wear-On, while BFDC aligns training with inference conditioning. In the MLLM-guided implementation, GBA supplies fine-grained garment features for detail preservation. Experiments on VITON-HD and DressCode support the combined strategy across diffusion-based and MLLM-guided backbones, showing improved image quality, more coherent limb layouts, and cleaner garment boundaries under substantial changes in garment silhouette and coverage.

{
    \small
    \bibliographystyle{ieeenat_fullname}
    \bibliography{main}
}

\clearpage
\setcounter{page}{1}
\maketitlesupplementary
\appendix

\section{Implementation Details.}
\label{sec:implementation_detail}
We follow the two-stage ToWo paradigm. The Body Structure Revealing (BSR) model is initialized from BooW-VTON~\citep{Zhang_2025_CVPRBOOW} and fine-tuned on the BF dataset using AdamW with a learning rate of $1.0 \times 10^{-5}$ and $(\beta_1,\beta_2)=(0.9,0.999)$. For the Wear-On stage, we fine-tune three backbones, namely TED-VITON, BooW-VTON, and an attention-modified OmniGen2, on the $\mathcal{D}_{\text{BFDC}}$ dataset with topology-consistent conditioning. TED-VITON and BooW-VTON follow their original training learning rates, while the OmniGen2 variant is optimized with AdamW using $(\beta_1,\beta_2)=(0.9,0.95)$ and a learning rate of $1.0 \times 10^{-6}$ for 10 epochs. All experiments are conducted on a single NVIDIA H100 GPU.

\section{Upper--Lower Difficulty Analysis}
\label{sec:Difficulty_Analysis}

\noindent Existing VTON methods exhibit a consistent performance gap between upper-body and lower-body synthesis, with lower-body results showing significantly higher KID values. This is an observed category-level difference under the reported evaluation protocol, rather than a controlled measurement of occlusion difficulty.
Table~\ref{tab:upper_lower_gap} reports the KID scores for upper- and lower-body synthesis on DressCode. We observe that baseline methods consistently present a substantial upper-lower gap. After integrating Take-Off and BFDC, this gap is reduced across multiple backbones.
Notably, for our model, the KID gap decreases from 0.119 (w/o ToWo) to 0.022, indicating improved consistency between upper- and lower-body synthesis. Here, the gap is the absolute difference between the two KID scores. The reduced gap describes more similar image-quality scores across categories; it does not isolate the cause of the original difference.

\begin{table}[t]
\centering
\footnotesize
\setlength{\tabcolsep}{4pt}
\renewcommand{\arraystretch}{1.18}

\caption{\textbf{Upper--lower KID gap on DressCode.}
ToWo reduces the gap across the evaluated backbones.
Gap denotes the absolute difference between lower- and upper-body KID.}
\label{tab:upper_lower_gap}

\begin{tabularx}{\linewidth}{l*{3}{>{\centering\arraybackslash}X}}
\toprule
\textbf{Method}
& \shortstack{\textbf{Lower-body}\\KID$\downarrow$}
& \shortstack{\textbf{Upper-body}\\KID$\downarrow$}
& \textbf{Gap} \\
\midrule
IDM-VTON~\citep{10.1007/978-3-031-73016-0_13}
& 0.497 & 0.304 & 0.193 \\
CAT-VTON~\citep{chong_catvton_2025}
& 1.012 & 0.645 & 0.367 \\
\midrule
TED-VITON~\citep{wan2025tedvitontransformerempowereddiffusionmodels}
& 0.762 & 0.373 & 0.389 \\
\rowcolor[HTML]{FFFFC7}
\quad + ToWo
& 0.584 & 0.278 & 0.306 \\
\midrule
BooW-VTON~\citep{Zhang_2025_CVPRBOOW}
& 0.509 & 0.289 & 0.220 \\
\rowcolor[HTML]{FFFFC7}
\quad + ToWo
& 0.376 & 0.220 & 0.156 \\
\midrule
Ours (w/o ToWo)
& 0.415 & 0.296 & 0.119 \\
\rowcolor[HTML]{FFFFC7}
\textbf{Ours}
& \textbf{0.172}
& \textbf{0.194}
& \textbf{0.022} \\
\bottomrule
\end{tabularx}
\end{table}

\section{In-the-Wild (ITW) Try-On}
\label{sec:ITW_Try-on}

We evaluate the complete ToWo-VTON pipeline on try-on examples with varied backgrounds. The model uses the same body-proxy and garment conditioning as in the standard setting. We assess whether it transfers the target garment while preserving the surrounding scene.

\subsection{Dress-Code ITW dataset curation}
Inspired by recent approaches using inpainting models for data augmentation~\citep{Wan_2025_ICCV}, we expand the in-the-wild domain coverage by leveraging the FLUX.1-Fill \citep{blackforestlabs_flux1fill_2024} inpainting model to replace the backgrounds of human images in the Dress-Code Lower-Body~\citep{10.1007/978-3-031-20074-8_20} dataset with diverse and realistic environments.
We generate contextual background prompts $P_{\text{context}}$ using GPT-5, which describe in detail common real-world scenarios for virtual try-on applications. Formally, the inpainting process is defined as:
\begin{equation}
X_{m\_bg} = \text{FLUX.1-Fill}\big(X_{m\_raw}, P_{\text{context}}\big),
\end{equation}
where $X_{m\_raw}$ denotes the background-removed human image obtained by erasing the background regions guided by the accompanying semantic map $X_{sm}$ provided in Dress-Code. This evaluates transfer to synthetically varied backgrounds. It does not provide independently generated garment transformations or a real-world capture benchmark, since the underlying person and garment-transformation protocol remain derived from DressCode.

\section{Tight-Fit Template Selection and Inference Policy}
\label{sec:tight_fit}

\begin{figure}[!ht]
     \centering
     \includegraphics[width=\linewidth]{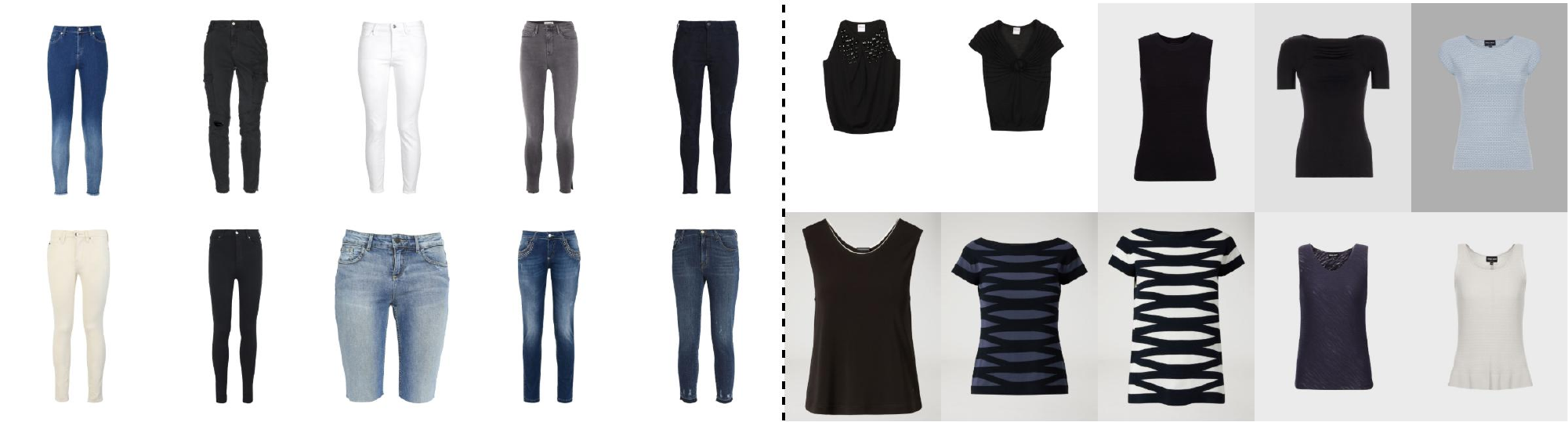}
     \caption{Visualization of Chosen Fit Templates.}
     \label{fig:dfit}

\end{figure}

For Stage~1 (Take-Off), we manually select 10 upper-body and 10 lower-body tight-fit garment templates, giving 20 templates in total (Fig.~\ref{fig:dfit}). The ten-template notation in the main text is per category, not the total template count or the BSR training-set size. Templates are chosen to expose body contours while keeping the person clothed.
During inference, one template is randomly sampled to generate $X^{bd}_m$. Wear-On also receives the target garment, so the proxy acts as a structural condition rather than the desired output appearance. Random sampling is the inference policy used in our experiments; it is not a controlled template-sensitivity study.
If the generated proxy is unsatisfactory, the separately trained direct Wear-On baseline can serve as a fallback by processing the original person image.

\section{User Study}
\label{sec: User_Study}

To assess perceptual quality, we conducted a user study on lower-body try-on results from DressCode. Each questionnaire contained 40 representative image sets, each displaying outputs from six methods: ToWo-VTON, SD-VITON, StableVITON, IDM-VTON, and reproduced TED-VITON and BooW-VTON. For each set, participants selected one preferred result for overall appearance and one for garment fitting. We received 107 questionnaire submissions and excluded seven incomplete submissions, retaining 100 completed questionnaires and 4,000 selections per criterion (8,000 in total).

\begin{figure}[!ht]
    \centering
    \includegraphics[width=1.0\linewidth]{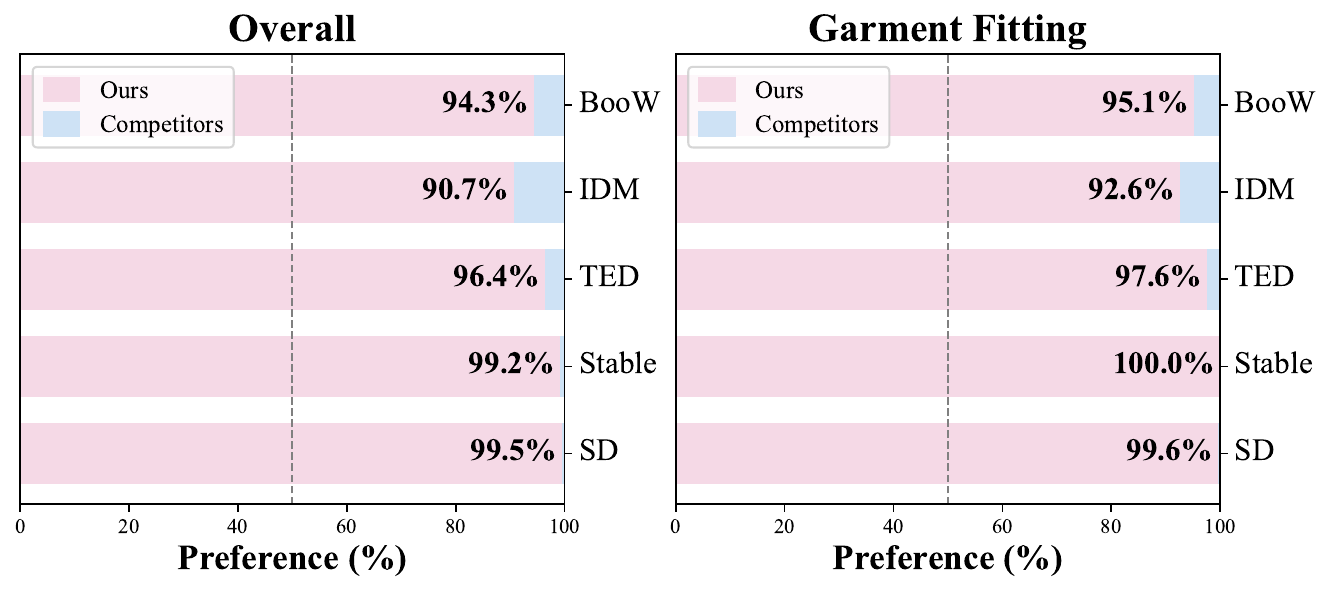}
    \caption{\textbf{User preferences on DressCode Lower-body.}}
       
    \label{fig:usd}
\end{figure}

For each criterion and baseline $b$, we summarize relative preference as $N_{\mathrm{ours}}/(N_{\mathrm{ours}}+N_b)$, where $N_{\mathrm{ours}}$ and $N_b$ denote the numbers of best-result selections received by ToWo-VTON and baseline $b$, respectively. Figure~\ref{fig:usd} presents these relative preference shares, derived from the six-method selection task rather than separate pairwise comparisons. These results characterize perceived quality on the 40 selected image sets and do not directly measure anatomical correctness or establish population-wide preference rates.

\begin{table}[t]
\centering
\footnotesize
\setlength{\tabcolsep}{3pt}
\renewcommand{\arraystretch}{1.18}

\caption{Quantitative results on the Dress-Code Lower-Body In-The-Wild dataset. \textbf{Bold} and \underline{underline} indicate the best and second-best results. FID is scaled by 100. * denotes reproduced implementations.}
\label{tab:ITW}

\begin{tabularx}{\linewidth}{l*{4}{>{\centering\arraybackslash}X}}
\specialrule{0.8pt}{0pt}{0pt}
\textbf{Dataset}
& \multicolumn{4}{c}{\textbf{Dress-Code ITW}} \\
\cmidrule(lr){2-5}
\textbf{Method}
& KID$\downarrow$
& FID$\downarrow$
& SSIM$\uparrow$
& LPIPS$\downarrow$ \\

\hline
\rowcolor[HTML]{EFEFEF}
\multicolumn{5}{c}{\textbf{Image Editing Methods}} \\
\hline
QWEN-Image \citep{wu2025qwenimagetechnicalreport}
& 2.012 & 32.440 & 0.422 & 0.532 \\
DreamOmni2 \citep{xia2025dreamomni2multimodalinstructionbasedediting}
& 1.173 & 26.264 & 0.519 & 0.412 \\

\hline
\rowcolor[HTML]{EFEFEF}
\multicolumn{5}{c}{\textbf{VTON Methods}} \\
\hline
SD-VITON \citep{10.1609/aaai.v38i5.28288}
& 8.060 & 87.053 & 0.581 & 0.458 \\
StableVITON \citep{Kim_2024_CVPR}
& 7.517 & 82.616 & 0.489 & 0.595 \\
IDM-VTON \citep{10.1007/978-3-031-73016-0_13}
& 0.562 & 17.838 & 0.689
& \cellcolor[HTML]{FFCE93}\underline{0.292} \\
CAT-VTON \citep{chong_catvton_2025}
& 0.760 & 20.376 & 0.675 & 0.313 \\
TED-VITON* \citep{wan2025tedvitontransformerempowereddiffusionmodels}
& 0.960 & 24.587 & 0.707 & 0.302 \\
BooW-VTON* \citep{Zhang_2025_CVPRBOOW}
& \cellcolor[HTML]{FFCE93}\underline{0.525}
& \cellcolor[HTML]{FFCE93}\underline{17.512}
& \cellcolor[HTML]{FFCE93}\underline{0.712}
& 0.297 \\

\hline
\textbf{Ours}
& \cellcolor[HTML]{FFFC9E}\textbf{0.333}
& \cellcolor[HTML]{FFFC9E}\textbf{16.046}
& \cellcolor[HTML]{FFFC9E}\textbf{0.722}
& \cellcolor[HTML]{FFFC9E}\textbf{0.285} \\
\specialrule{0.8pt}{0pt}{0pt}
\end{tabularx}
\end{table}

\section{Results on ITW Try-On}
On Dress-Code ITW, ToWo-VTON achieves a KID of 0.333 and an FID of 16.046, compared with 0.525 and 17.512 for the strongest competing baseline on these metrics (Table~\ref{tab:ITW}). It also achieves the highest SSIM (0.722) and lowest LPIPS (0.285) among the methods evaluated. These results characterize the full pipeline on the modified backgrounds.

\section{Computational Overhead}
\label{sec:Computational_Overhead}

Table~\ref{tab:timecross} compares the inference efficiency of ToWo-VTON with representative image editing and VTON baselines on an NVIDIA H100, while Table~\ref{tab:timecomp} further breaks down the component-wise cost of the MLLM-based Wear-On stage. 
Since Stage~1 is implemented as a BSR variant fine-tuned from BooW-VTON, it shares the same backbone and therefore has the same inference cost as BooW-VTON. 
The additional computational cost of ToWo-VTON mainly comes from Stage~2, where the MLLM-based DiT denoiser performs iterative garment synthesis.

Within Stage~2, the Qwen2.5-VL encoder is called only once per inference, while the DiT denoiser dominates the runtime and FLOPs. 
Over 50 steps, OmniGen2's native input refiners account for 4.332\,s and 1018.88\,T FLOPs, equivalent to 6.5\% of the main DiT stack's runtime and 3.3\% of its FLOPs. We report this portion of the backbone computation separately in Table~\ref{tab:timecomp}.
Overall, ToWo-VTON is more computationally demanding than single-stage VTON baselines, but remains more efficient than general-purpose MLLM-based image editing methods in both latency and memory. 
This provides a practical trade-off between efficiency and topology-aware try-on quality.

\begin{table}[t]
\centering
\footnotesize
\setlength{\tabcolsep}{2.5pt}
\renewcommand{\arraystretch}{1.18}

\caption{Inference efficiency comparison with image editing and VTON baselines. Stage~1 is the BSR variant fine-tuned from BooW-VTON and has the same inference cost as BooW-VTON.}
\label{tab:timecross}

\begin{tabularx}{\linewidth}{
    >{\raggedright\arraybackslash}p{0.38\linewidth}
    *{3}{>{\centering\arraybackslash}X}
}
\specialrule{0.8pt}{0pt}{0pt}
\textbf{Method}
& \shortstack{\textbf{Time}\\\textbf{(s)}}
& \shortstack{\textbf{Peak VRAM}\\\textbf{(GB)}}
& \shortstack{\textbf{FLOPs}\\\textbf{(T)}} \\

\hline
\rowcolor[HTML]{EFEFEF}
\multicolumn{4}{c}{\textbf{MLLM-Based Methods}} \\
\hline
QWEN-Image
& 546.50 & 59.96 & 26121.59 \\
DreamOmni2
& 312.253 & 53.97 & 14367.07 \\

\hline
\rowcolor[HTML]{EFEFEF}
\multicolumn{4}{c}{\textbf{VTON Methods}} \\
\hline
IDM-VTON
& 11.911 & 13.113 & 694.09 \\
BooW-VTON
& 10.359 & 10.98 & 1221.8 \\

\hline
\rowcolor[HTML]{EFEFEF}
\multicolumn{4}{c}{\textbf{Ours}} \\
\hline
Shared Stage~1
& 10.359 & 10.98 & 1221.8 \\
Stage~2 (Diffusion-based)
& 10.359 & 10.98 & 1221.8 \\
Stage~2 (MLLM-based)
& 73.789 & 27.44 & 31848.8 \\
\specialrule{0.8pt}{0pt}{0pt}
\end{tabularx}
\end{table}

\begin{table}[t]
\centering
\footnotesize
\setlength{\tabcolsep}{2.5pt}
\renewcommand{\arraystretch}{1.18}

\caption{Component-wise cost of the MLLM-based Wear-On stage over 50 denoising steps. The input refiners are part of the OmniGen2 backbone.}
\label{tab:timecomp}

\begin{tabularx}{\linewidth}{
    >{\raggedright\arraybackslash}p{0.38\linewidth}
    *{3}{>{\centering\arraybackslash}X}
}
\toprule
\textbf{Component}
& \shortstack{\textbf{Time}\\\textbf{(s)}}
& \shortstack{\textbf{Peak VRAM}\\\textbf{(GB)}}
& \shortstack{\textbf{FLOPs}\\\textbf{(T)}} \\
\midrule
Qwen2.5-VL Encoder
& 2.827 & 22.56 & 1.46 \\
\addlinespace[2pt]
Backbone input refiners (50 steps)
& 4.332 & 26.25 & 1018.88 \\
\addlinespace[2pt]
DiT (50 steps)
& 66.821 & 25.96 & 30823.04 \\
\bottomrule
\end{tabularx}
\end{table}

\section{Fairness of Unpaired Evaluation}
\label{sec:Fairness_unpair}
Paired evaluation reconstructs the original person image, but its input construction depends on the model interface. Mask-based methods mask the garment region and reconstruct it conditioned on the original garment. Mask-free methods instead receive a synthetically re-dressed person and the original garment, and regenerate the full original image. The latter therefore requires full-image reconstruction from an altered source rather than masked-region reconstruction. Paired SSIM/LPIPS should be interpreted with this difference in mind. In unpaired evaluation, both interfaces synthesize a target garment different from the source clothing; KID/FID assess the generated image distribution. Reported FID is multiplied by 100. These settings need not match the protocols of the original baseline publications.
BooW-VTON is used both to create BSR training inputs and to construct benchmark transformations. These uses involve separate training and test splits, but the shared generator may introduce distribution-specific advantages. Cross-backbone improvements establish applicability to the tested architectures, not independence from the augmentation generator. The background-varied ITW experiment probes a different source of variation and does not remove this limitation.

\section{Additional Qualitative Evaluation}
\label{sec:Additional_qualitative}
\begin{figure*}[t]
     \centering
     \includegraphics[width=\textwidth]{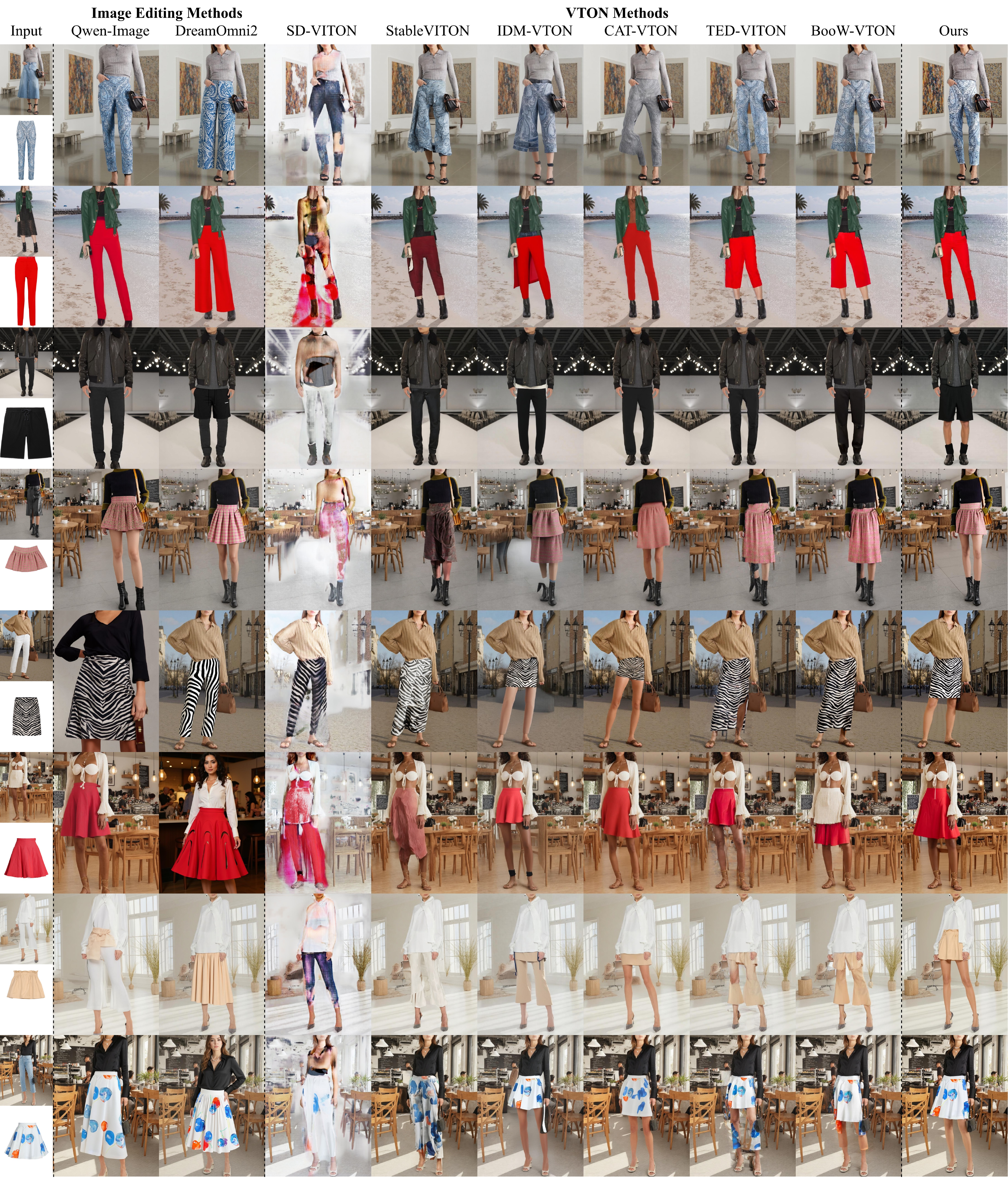}
     \caption{In-the-Wild Lower-Body Try-On Comparison. Despite being trained on the standard DressCode lower-body without ITW fine-tuning, our method demonstrates superior robustness to background variation and better preservation of garment and identity details.}
     \label{fig:vd-itw}
\end{figure*}

\begin{figure*}[t]
     \centering
     \includegraphics[width=0.9\textwidth]{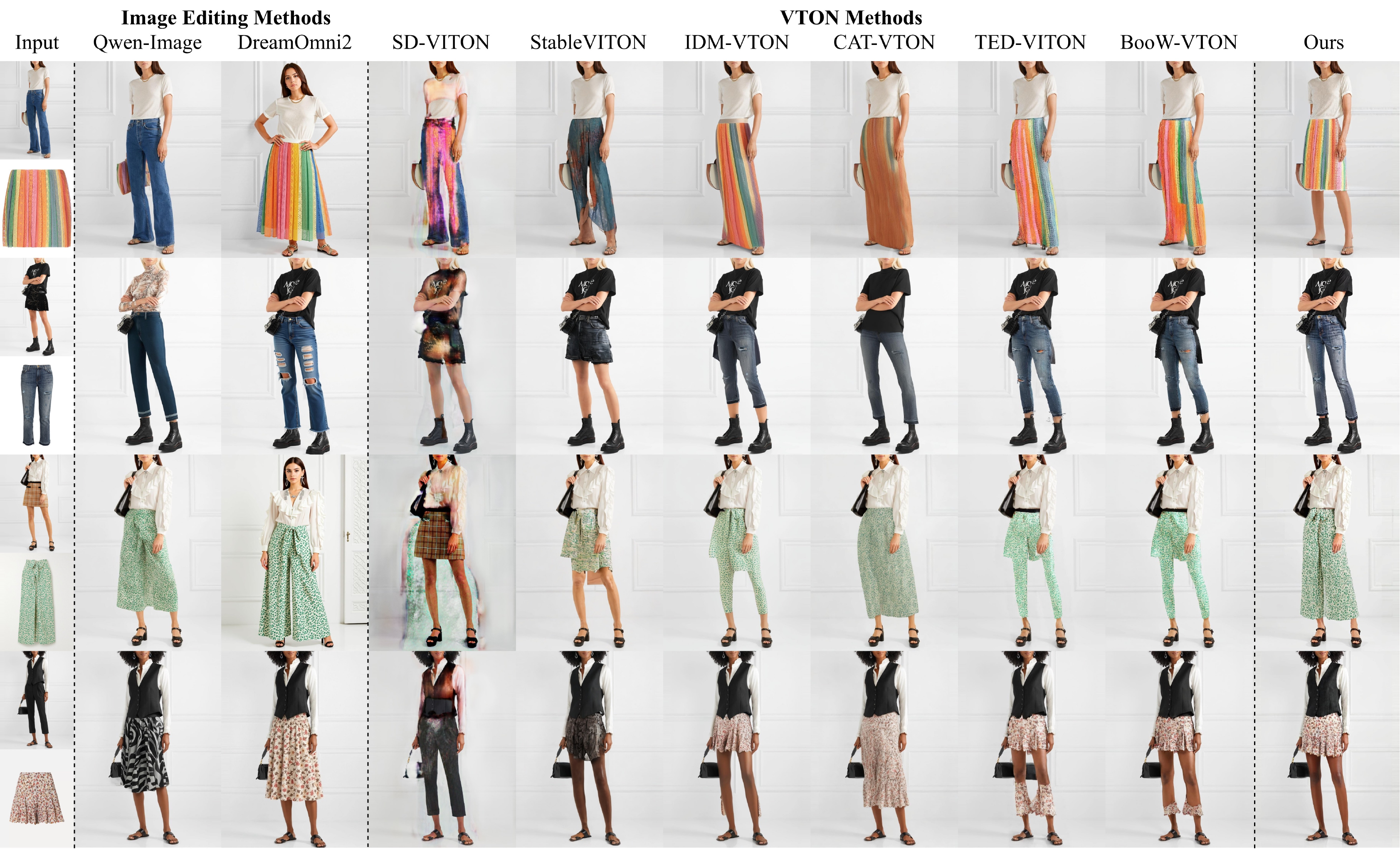}
     \caption{Additional Qualitative Evaluation on Dress-Code Lower-Body dataset. ToWo-VTON achieves significantly more consistent body alignment, human identity retention, and superior preservation of garment spatial details compared to existing approaches.}
     \label{fig:vd-LBCOMP}

\end{figure*}

\begin{figure*}[t]
     \centering
     \includegraphics[width=0.9\textwidth]{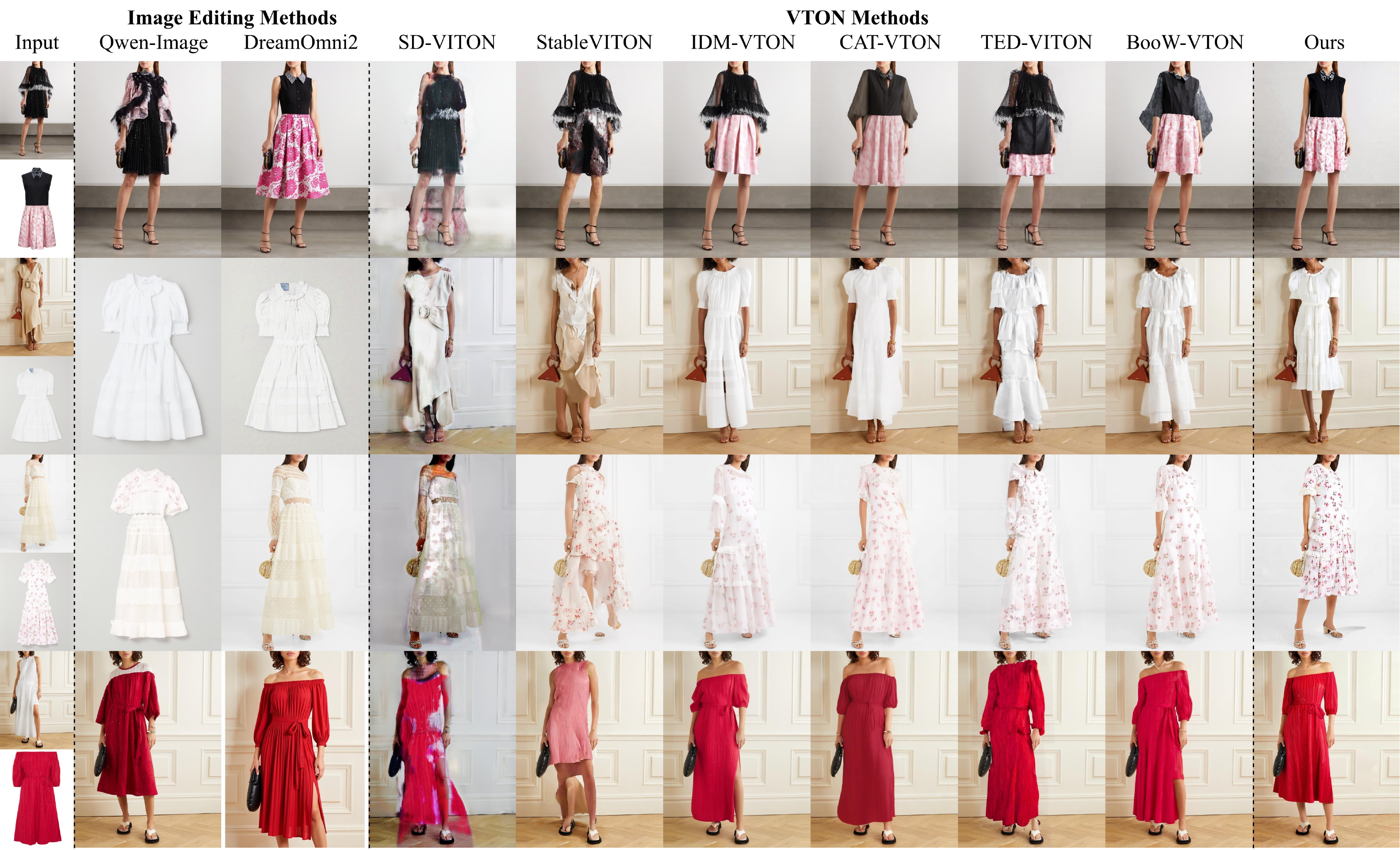}
     \caption{Additional Qualitative Evaluation on Dress-Code Dresses dataset. ToWo-VTON achieves significantly more consistent body alignment, human identity retention, and superior preservation of garment spatial details compared to existing approaches.}
     \label{fig:vd-DRCOMP}
\end{figure*}

\begin{figure*}[t]
     \centering
     \includegraphics[width=0.9\textwidth]{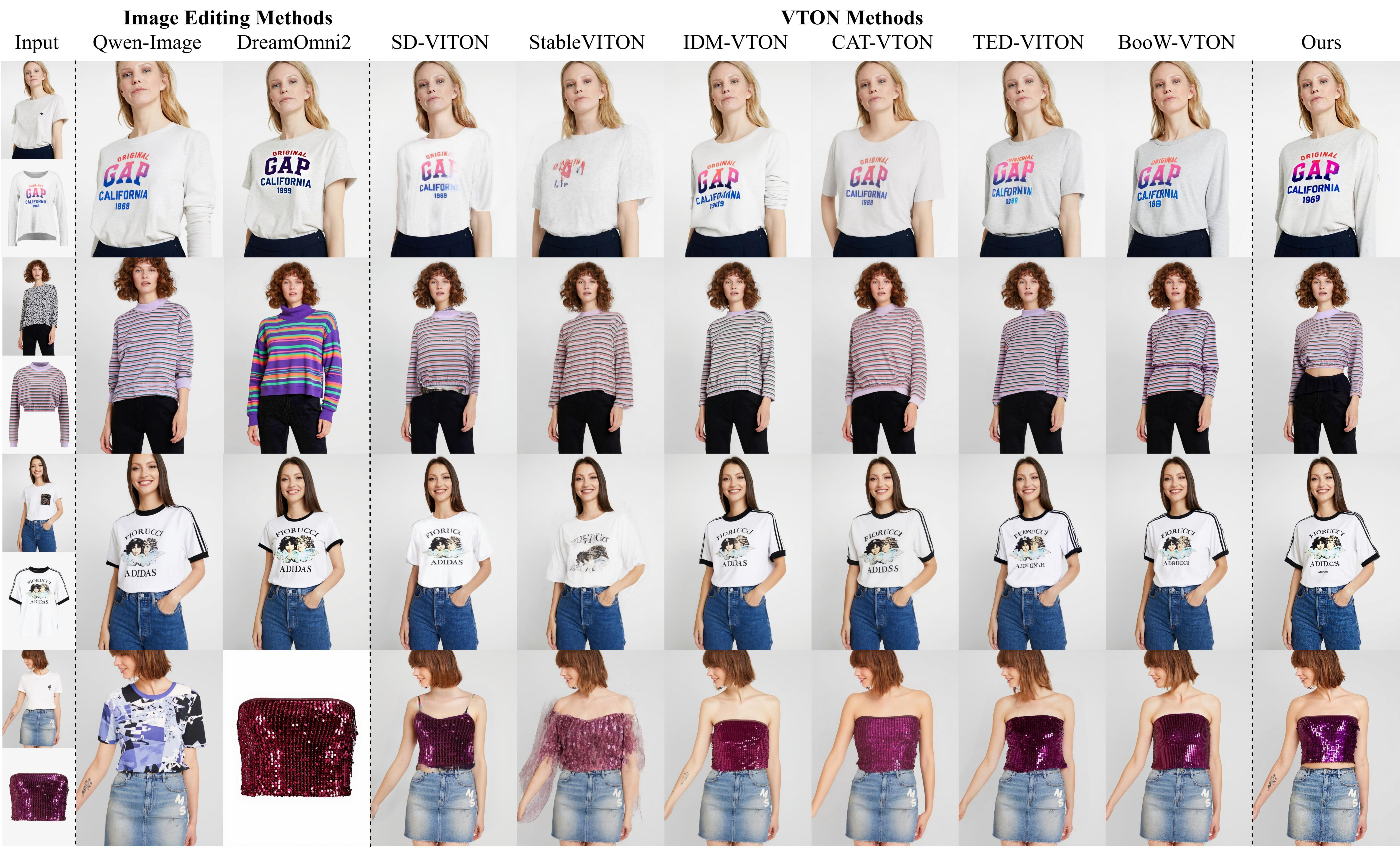}
     \caption{Qualitative comparison on VITON-HD dataset. ToWo-VTON achieves more consistent garment–body alignment and clearer garment details than prior methods.}
     \label{fig:vd-VITCOMP}
\end{figure*}

\begin{figure*}[t]
     \centering
     \includegraphics[width=0.9\textwidth]{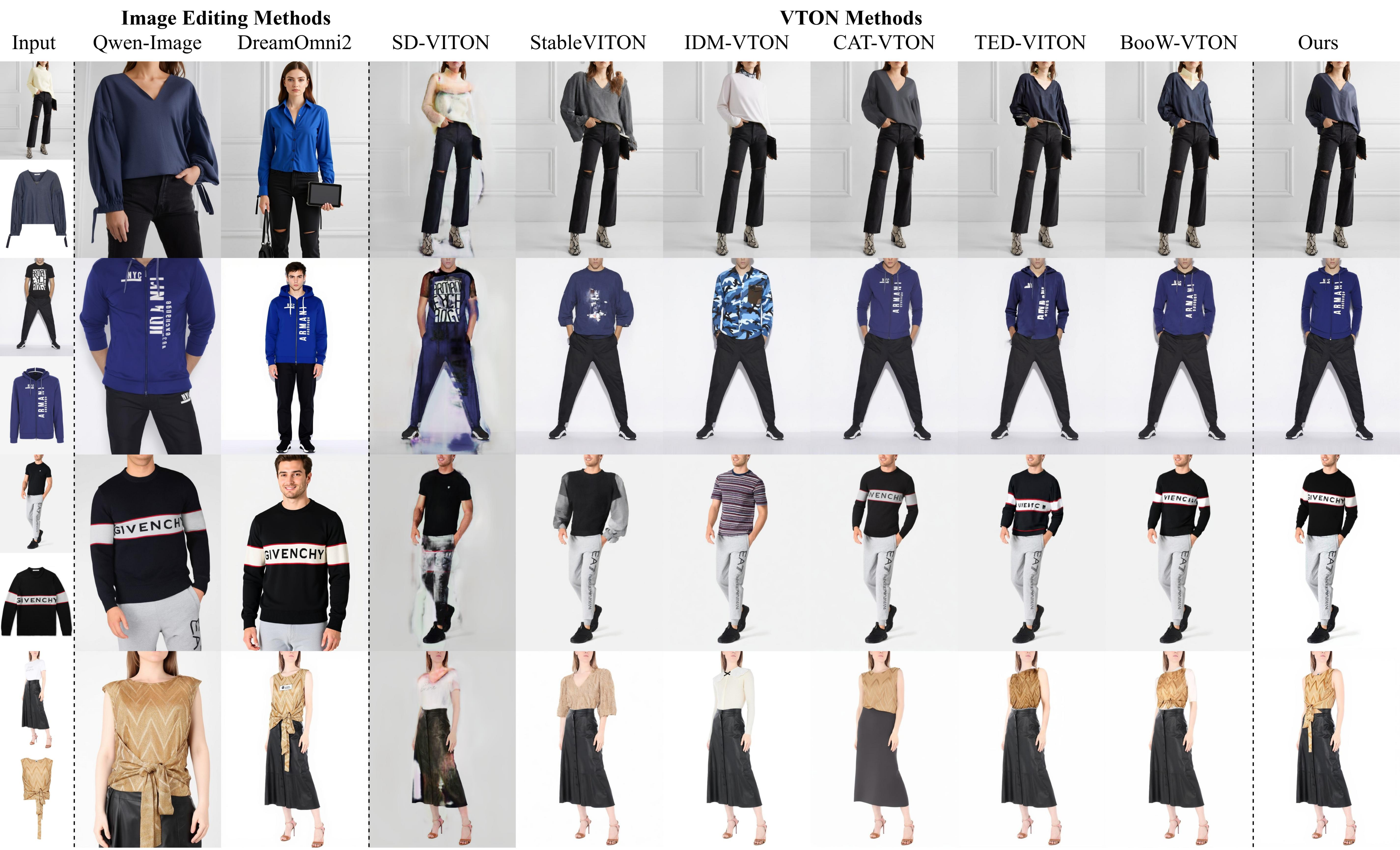}
     \caption{Qualitative comparison on Dress-Code Upper-Body dataset. ToWo-VTON achieves more consistent garment–body alignment and clearer garment details than prior methods.}
     \label{fig:Dress_code_upper_comparison}
\end{figure*}

We provide additional qualitative results to further evaluate ToWo-VTON across upper-body, lower-body, and full-body garment categories on DressCode and VITON-HD. 
As shown in Fig.~\ref{fig:vd-itw}, our method produces the shown lower-body results on synthetic background variations without ITW fine-tuning. 
The comparisons in Fig.~\ref{fig:vd-LBCOMP} and Fig.~\ref{fig:vd-DRCOMP} further demonstrate that ToWo-VTON produces more stable body alignment, better identity preservation, and more faithful garment fitting under challenging lower-body and dress try-on scenarios. 
Additional upper-body results on VITON-HD and DressCode are provided in Fig.~\ref{fig:vd-VITCOMP} and Fig.~\ref{fig:Dress_code_upper_comparison}, where our method maintains consistent garment--body alignment while preserving fine-grained garment details. 
These examples illustrate garment fitting and visible body layout across the displayed cases; they do not provide performance stratified by occlusion severity or transformation magnitude.


\begin{figure*}[t]
     \centering
     \includegraphics[width=0.9\textwidth]{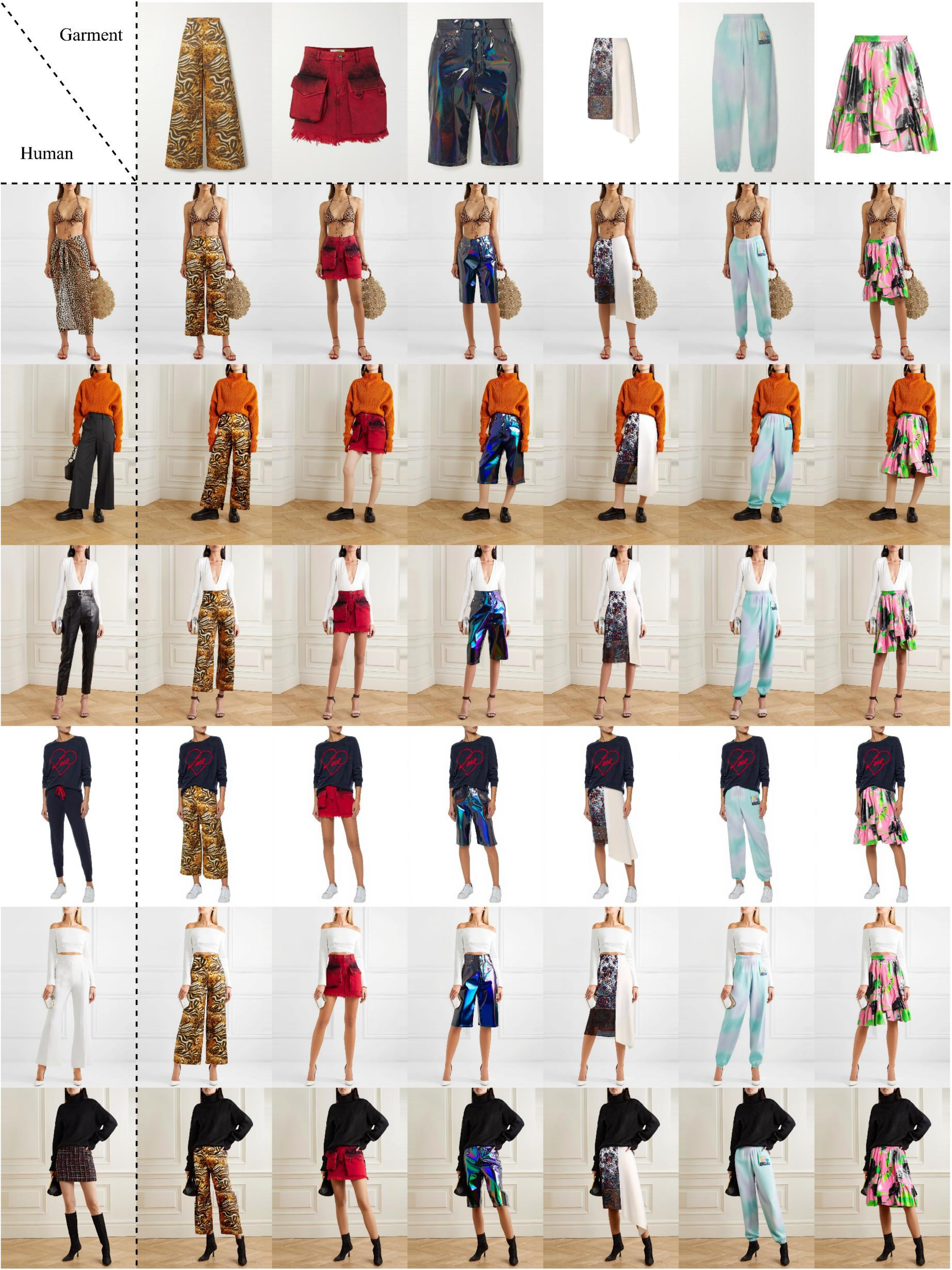}
     \caption{ToWo-VTON DressCode Lower-Body Try-on Results. Visual demonstration of garment transfer robustness across diverse poses and clothing types.}
     \label{fig:vd-lb}

\end{figure*}

\begin{figure*}[t]
     \centering
     \includegraphics[width=0.9\textwidth]{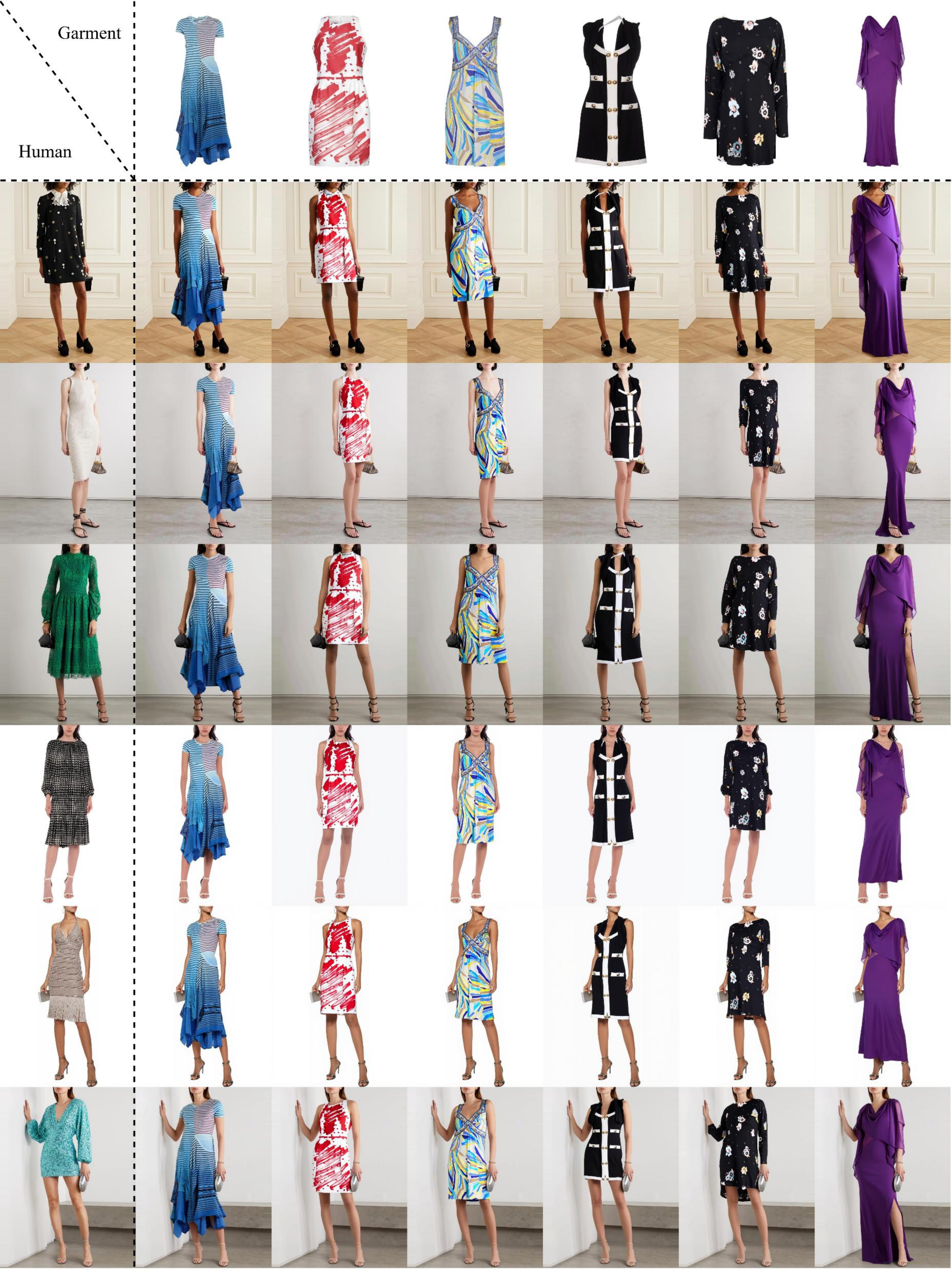}
     \caption{ToWo-VTON DressCode Full-Body (Dresses) Try-on Results. Visual demonstration of garment transfer robustness across diverse poses and clothing types.}
     \label{fig:vd-dr}

\end{figure*}

\begin{figure*}[t]
     \centering
     \includegraphics[width=0.9\textwidth]{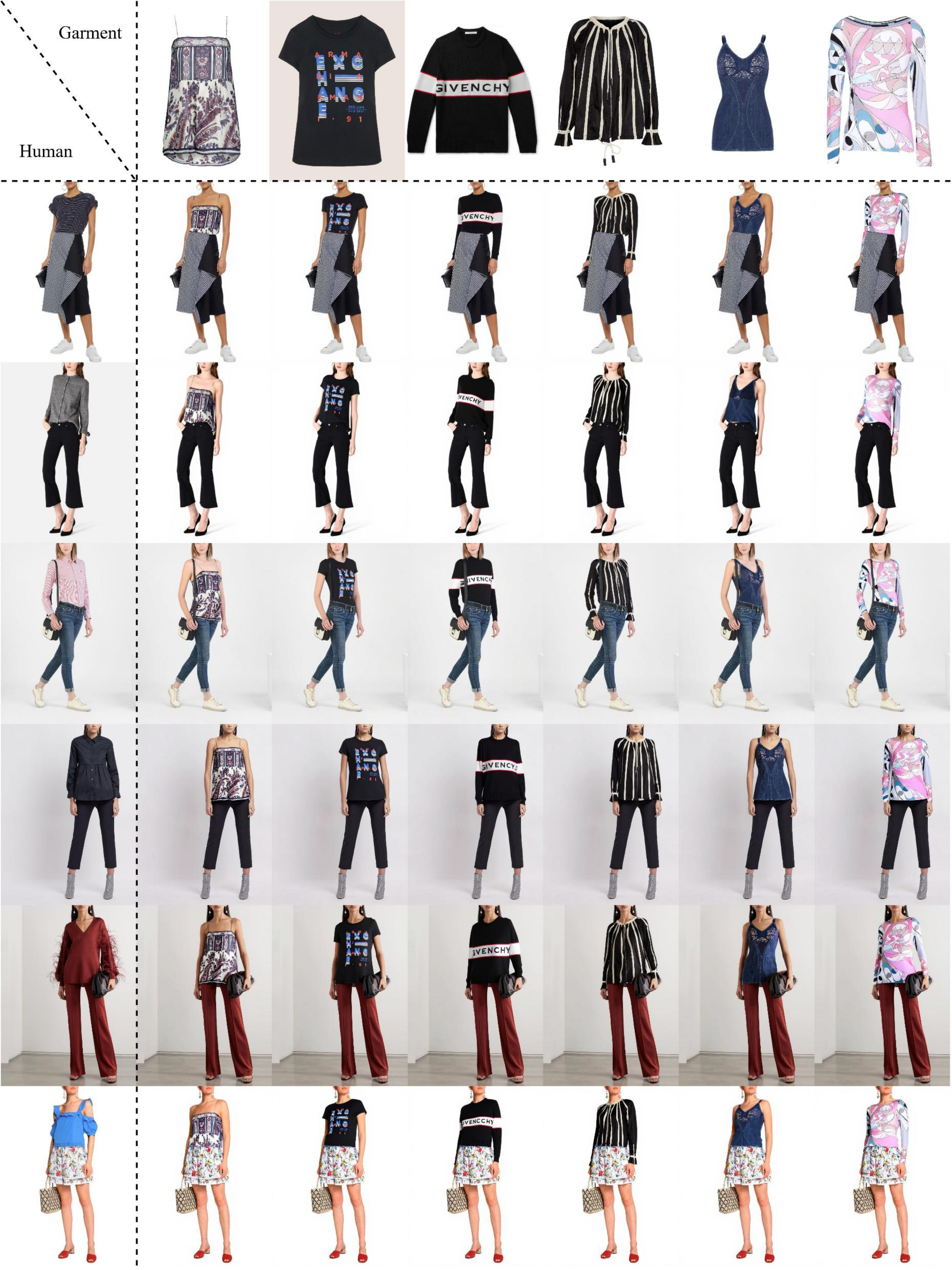}
     \caption{ToWo-VTON DressCode Upper-Body Try-on Results. Visual demonstration of garment transfer robustness across diverse poses and clothing types.}
     \label{fig:vd-ub}

\end{figure*}

\begin{figure*}[t]
     \centering
     \includegraphics[width=0.9\textwidth]{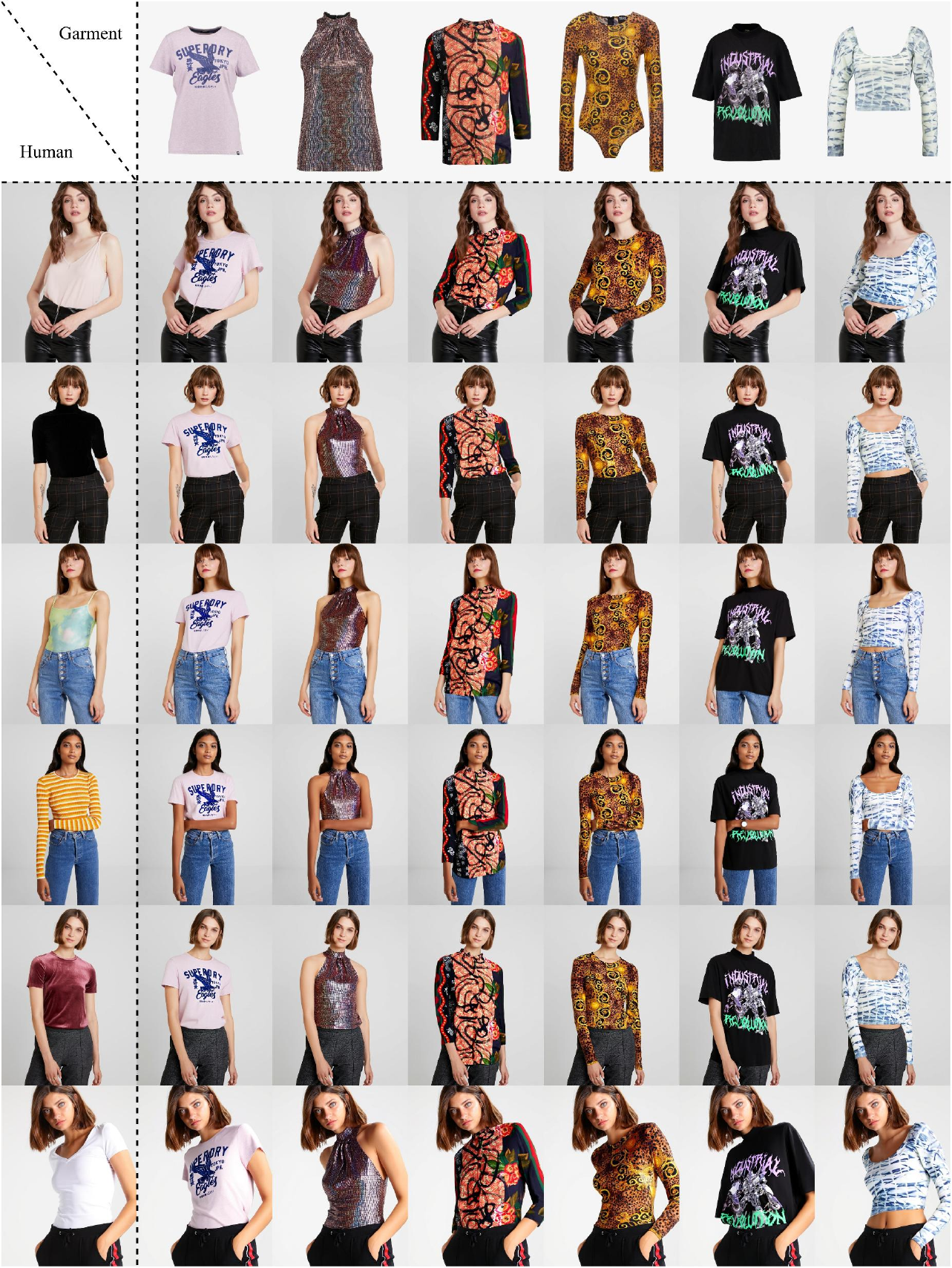}
     \caption{ToWo-VTON VITON-HD Try-on Results. Visual demonstration of garment transfer robustness across diverse poses and clothing types.}
     \label{fig:vd-viton}

\end{figure*}
\clearpage

\end{document}